\documentclass[10pt,twocolumn,letterpaper]{article}
\usepackage[pagenumbers]{cvpr}
\usepackage{graphicx}
\usepackage{amsmath}
\usepackage{amssymb}
\usepackage{booktabs}
\usepackage{caption}
\usepackage{cuted}
\usepackage{float}
\usepackage{graphicx}
\usepackage{pifont}
\usepackage{xcolor}
\usepackage{listings}
\usepackage{bm}
\usepackage[square,numbers,sort]{natbib}
\usepackage[pagebackref,breaklinks,colorlinks]{hyperref}

\usepackage[capitalize]{cleveref}
\crefname{section}{Sec.}{Secs.}
\Crefname{section}{Section}{Sections}
\Crefname{table}{Table}{Tables}
\crefname{table}{Tab.}{Tabs.}

\makeatletter
\renewcommand{\paragraph}{%
  \@startsection{paragraph}{4}%
  {\z@}{0.25em}{-1em}%
  {\normalfont\normalsize\bfseries}}
\makeatother

\newcommand{\cmark}{\ding{51}}

\newcommand{\methodname}{RealFusion\xspace}
\newcommand{\Methodname}{RealFusion\xspace}
\newcommand{\dreamfusion}{DreamFusion\xspace}
\newcommand{\etoken}{$\langle\textbf{e}\rangle$}
\newcommand{\colorhref}[3][blue]{\href{#2}{\color{#1}{#3}}}%

\usepackage{amsmath,amsfonts,bm}

\def\eqref#1{equation~\ref{#1}}

\def\1{\bm{1}}

\def\eps{{\epsilon}}

\DeclareMathAlphabet{\mathsfit}{\encodingdefault}{\sfdefault}{m}{sl}
\SetMathAlphabet{\mathsfit}{bold}{\encodingdefault}{\sfdefault}{bx}{n}

\newcommand{\degrees}{^{\circ}}

\def\cvprPaperID{9790} %
\def\confName{CVPR}
\def\confYear{2023}

\title{\methodname\\ 360$^\circ$ Reconstruction of Any Object from a Single Image}

\author{Luke Melas-Kyriazi
\quad Iro Laina
\quad Christian Rupprecht
\quad Andrea Vedaldi
\\
\\
\small Visual Geometry Group, Department of Engineering Science, University of Oxford\\
\small {\texttt{\{lukemk,iro,chrisr,vedaldi\}@robots.ox.ac.uk}}\\
\small {\colorhref{https://lukemelas.github.io/realfusion}{https://lukemelas.github.io/realfusion}}
}

\begin{document}
\maketitle
\begin{strip}
\vspace{-3em}
\centering
\includegraphics[width=\linewidth]{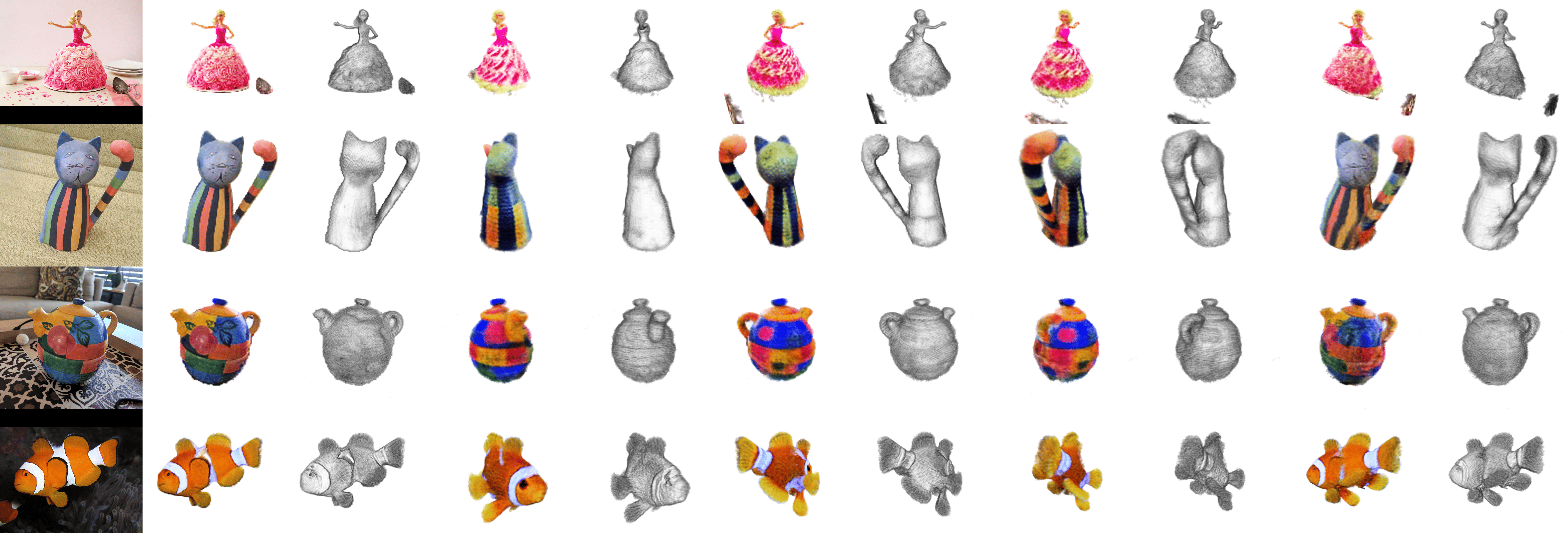}
\vspace{-1em}
\captionof{figure}{
\textbf{\Methodname} generates a full 360$^\circ$ reconstruction of any object given a \emph{single image} of it (left column).
It does so by leveraging an existing diffusion-based 2D image generator.
From the given image, it synthesizes a prompt that causes the diffusion model to ``dream up'' other views of the object.
It then extracts a neural radiance field from the original image and the diffusion model-based prior, thereby reconstructing the object in full.
Both appearance and geometry are reconstructed faithfully and extrapolated in a plausible manner (see the textured and shaded reconstructions from different viewpoints).}
\label{f:splash}
\end{strip}
\begin{abstract}
We consider the problem of reconstructing a full 360$^\circ$ photographic model of an object from a single image of it.
We do so by fitting a neural radiance field to the image, but find this problem to be severely ill-posed.
We thus take an off-the-self conditional image generator based on diffusion and engineer a prompt that encourages it to ``dream up'' novel views of the object.
Using the recent \dreamfusion method, we fuse the given input view, the conditional prior, and other regularizers in a final, consistent reconstruction.
We demonstrate state-of-the-art reconstruction results on benchmark images when compared to prior methods for monocular 3D reconstruction of objects.
Qualitatively, our reconstructions provide a faithful match of the input view and a plausible extrapolation of its appearance and 3D shape, including to the side of the object not visible in the image.
\end{abstract}
\section{Introduction}%
\label{s:intro}

We consider the problem of obtaining a 360$^\circ$ photographic reconstruction of \emph{any} object given a \emph{single image} of it.
The challenge is that a single image \emph{does not} contain sufficient information for 3D reconstruction.
Without access to multiple views, an image only provides weak evidence about the 3D shape of the object, and only for one side of it.
Even so, there is proof that this task \emph{can} be solved:
any skilled 3D artist can take a picture of almost any object and, given sufficient time and effort, create a plausible 3D model of it.
The artist can do so by tapping into her vast knowledge of the natural world and of the objects it contains, making up for the information missing in the image.

To solve this problem algorithmically, one must then marry visual geometry with a powerful statistical model of the 3D world.
The recent explosion of 2D image generators like DALL-E~\cite{ramesh2021zero}, Imagen~\cite{saharia2022photorealistic}, and Stable Diffusion~\cite{rombach2022high} suggests that such models might not be far behind.
By using diffusion, these methods can solve highly-ambiguous generation tasks, obtaining plausible 2D images from textual descriptions, semantic maps, partially-complete images, or simply unconditionally from random noise.
Clearly, these models possess high-quality priors---if not of the 3D world, then at least of the way it is represented in 2D images.
Hence, in theory, a 3D diffusion model trained on vast quantities of 3D data should be capable of producing 3D reconstructions, either unconditionally or conditioned on a 2D image.
However, training such a model is infeasible because, while one can access billions of 2D images~\cite{schuhmann2022laion}, the same cannot be said about 3D data.

The alternative to training a 3D diffusion model is to extract 3D information from an existing 2D model.
A 2D image generator can in fact be used to sample or validate multiple views of a given object;
these multiple views can then be used to perform 3D reconstruction.
With early GAN-based generators, authors showed some success for simple data like faces and synthetic objects~\cite{gadelha163d-shape,wu16learning,henzler19escaping,nguyen-phuoc19hologan:,nguyen-phuoc20blockgan:,chan22efficient}.
With the availability of large-scale models like CLIP~\cite{radford2021learning} and, more recently, diffusion models, increasingly complex results have been obtained.
The most recent example is \dreamfusion~\cite{poole22dreamfusion:}, which generates high-quality 3D models from textual descriptions alone.

Despite these advances, the problem of single-image 3D reconstruction remains largely unsolved. In fact, these recent methods do not solve this problem. %
They either sample random objects, or, like in the case of \dreamfusion, start from a textual description.

A problem in extending generators to reconstruction is \emph{coverage} (sometimes known as mode collapse).
For example, high-quality face generators based on GANs are usually difficult to invert:
they may be able to generate \emph{many} different high-quality images, and yet are usually unable to generate \emph{most} images~\cite{bau19seeing}.
Conditioning on an image provides a much more detailed and nuanced specification of the object than, say, a textual description.
It is not obvious if the generator model would be able to satisfy all such constraints.

In this paper, we study this problem in the context of diffusion models.
We express the object's 3D geometry and appearance by means of a neural radiance field.
Then, we train the radiance field to reconstruct the given input image by minimizing the usual rendering loss.
At the same time, we sample random other views of the object, and constrain them with the diffusion prior, using a technique similar to \dreamfusion.

We find that, out of the box, this idea does not work well.
Instead, we need to make a number of improvements and modifications.
The most important change is to adequately condition the diffusion model.
The idea is to configure the prior to ``dream up'' or sample images that may \emph{plausibly constitute other views of the given object}.
We do so by engineering the diffusion prompt from random augmentations of the given image.
Only in this manner does the diffusion model provide sufficiently strong constraints to allow meaningful 3D reconstruction.

In addition to setting the prompt correctly, we also add some regularizers:
shading the underlying geometry and randomly dropping out texture (also similar to \dreamfusion),
smoothing the normals of the surface, 
and fitting the model in a coarse-to-fine fashion, capturing first the overall structure of the object and only then the fine-grained details.
We also focus on efficiency and base our model on InstantNGP~\cite{mueller2022instant}.
In this manner, we achieve reconstructions in the span of hours instead of days if we were to adopt traditional MLP-based NeRF models.

We assess our approach by using random images captured in the wild as well as existing benchmark datasets.
Note that we do \emph{not} train a fully-fledged 2D-to-3D model and we are \textit{not} limited to specific object categories; rather, we perform reconstruction on an image-by-image basis using a pretrained 2D generator as a prior. 
Nonetheless, we can surpass quantitatively and qualitatively previous single-image reconstructors, including Shelf-Supervised Mesh Prediction~\cite{ye21shelf-supervised}, which uses supervision tailored specifically for 3D reconstruction.

More impressively, and more importantly, we obtain plausible 3D reconstructions that are a good match for the provided input image (\cref{f:splash}).
Our reconstructions are not perfect, as the diffusion prior clearly does its best to explain the available image evidence but cannot always match all the details.
Even so, we believe that our results convincingly demonstrate the viability of this approach and trace a path for future improvements.

To summarize, we make the following \textbf{contributions}:
(1) We propose \methodname, a method that can extract from a single image of an object a 360$^\circ$ photographic 3D reconstruction without assumptions on the type of object imaged or 3D supervision of any kind;
(2) We do so by leveraging an existing 2D diffusion image generator via a new single-image variant of textual inversion;
(3) We also introduce new regularizers and provide an efficient implementation using InstantNGP\@;
(4) We demonstrate state-of-the-art reconstruction results on a number of in-the-wild images and images from existing datasets when compared to alternative approaches.

\section{Related work}\label{s:related}

\paragraph{Image-based reconstruction of appearnce and geometry.}

Much of the early work on 3D reconstruction is based on principles of multi-view geometry~\cite{Hartley2004}.
These classic methods use photometry only to match image features and then discard it and only estimate 3D shape.

The problem of reconstructing photometry and geometry together has been dramatically revitalized by the introduction of neural radiance fields (RFs). %
NeRF~\cite{mildenhall20nerf:} in particular noticed that a coordinate MLP provides a compact and yet expressive representation of 3D fields, and can be used to model RFs with great effectiveness.
Many variants of NeRF-like models have since appeared.
For instance, some~\cite{wang21neus:,ueda22neural,lin20sdf-srn:} use sign distance functions (SDFs) to recover cleaner geometry.
These approaches assume that dozens if not hundreds of views of each scene are available for reconstruction.
Here, we use them for single-image reconstruction, using a diffusion model to ``dream up'' the missing views.

\paragraph{Few-view reconstruction.}

Many authors have attempted to improve the statistical efficiency of NeRF-like models, by learning or incorporating various kinds of priors.
Quite related to our work, NeRF-on-a-Diet~\cite{jain21putting} reduces the number of images required to learn a NeRF by generating random views and measuring their ``semantic compatibility'' with the available views via CLIP embeddings~\cite{radford21learning}, but they still require several input views.

While CLIP is a general-purpose model learned on 2D data, other authors have learned deep networks specifically for the goal of inferring NeRFs from a small number of views.
Examples include
IBRNet~\cite{wang21ibrnet:},
NeRF-WCE~\cite{henzler21unsupervised},
PixelNeRF~\cite{yu2021pixelnerf},
NeRFormer~\cite{reizenstein21co3d},
and ViewFormer~\cite{kulhanek22viewformer:}.
These models still generally require more than one input view at test time, require multi-view data for training, and are often optimized for specific object categories.

\paragraph{Single-view reconstruction.}

Some authors have attempted to recover full radiance fields from single images, but this generally requires multi-view data for training, as well as learning models that are specific to a specific object category.
3D-R2N2~\cite{choy163d-r2n2:},
Pix2Vox~\cite{xie20pix2vox:,xie20pix2vox:},
and
LegoFormer~\cite{yagubbayli21legoformer:} learn to reconstruct volumetric representation of simple objects, mainly from synthetic data like ShapeNet~\cite{chang15shapenet}.
More recently, CodeNeRF~\cite{jang21codenerf:} predicts a full radiance field, including reconstructing the photometry of the objects.
AutoRF~\cite{muller22autorf:} learns a similar autoencoder specifically for cars.

\paragraph{Extracting 3D models from 2D generators.}

Several authors have proposed to extract 3D models from 2D image generators, originally using GANs~\cite{gadelha163d-shape,wu16learning,henzler19escaping,nguyen-phuoc19hologan:,nguyen-phuoc20blockgan:,chan22efficient}.

More related to our work, CLIP-Mesh~\cite{khalidclip-mesh:} and Dream Fields~\cite{jain22zero-shot} do so by using the CLIP embedding and can condition 3D generation on text.
Our model is built on the recent Dream Fusion approach~\cite{poole22dreamfusion:}, which builds on a similar idea using a diffusion model as prior.

However, these models have been used as either pure generators or generators conditioned on vague cues such as class identity or text.
Here, we build on similar ideas, but we apply them to the case of single-view reconstruction.

Recently, the authors of~\cite{watson22novel} have proposed to directly generate multiple 2D views of an object, which can then be reconstructed in 3D using a NeRF-like model.
This is also reminiscent of our approach, but their model requires multi-view data for training, is only tested on synthetic data, and requires to explicitly sample multiple views for reconstruction (in our case they remain implicit).

\paragraph{Diffusion Models.}

Diffusion denoising probabilistic models are a class of generative models based on iteratively reversing a Markovian noising process.
In vision, early works formulated the problem as learning a variational lower bound~\cite{ho20denoising}, or framed it as optimizing a score-based generative model~\cite{song19generative,song2020improved} or as the discretization of a continuous stochastic process~\cite{song21score-based}.
Recent improvements includes the use of faster and deterministic sampling~\cite{liu2021pseudo,watson2021learning,ho20denoising}, class-conditional models~\cite{dhariwal2021diffusion,song2020improved}, text-conditional models~\cite{nichol2021glide}, and modeling in latent space~\cite{rombach22high-resolution}.

\begin{figure}[t]
\centering
\includegraphics[width=1.0\linewidth]{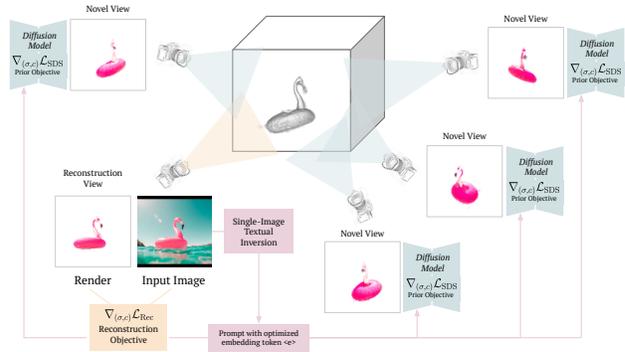}
\caption{
    \textbf{Method diagram.} Our method optimizes a neural radiance field using two objectives simultaneously: a reconstruction objective and a prior objective. The reconstruction objective ensures that the radiance field resembles the input image from a specific, fixed view. The prior objective uses a large pre-trained diffusion model to ensure that the radiance field looks like the given object from randomly sampled novel viewpoints. The key to making this process work well is to condition the diffusion model on a prompt with a custom token \etoken, which is generated prior to reconstruction using single-image textual inversion. This diagram does not display our coarse-to-fine training strategy or regularization terms, both of which improve qualitative results.
}
\label{fig:method}
\end{figure}

\begin{figure}[t]
\centering
\includegraphics[width=1.0\linewidth]{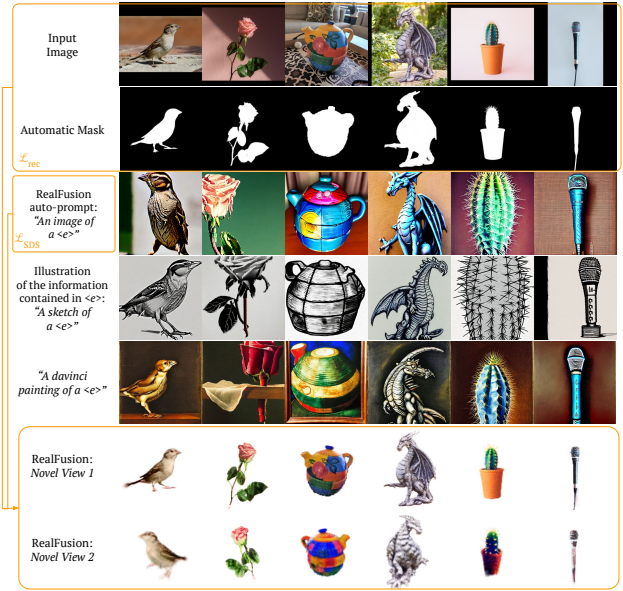}
\caption{
    \textbf{Examples demonstrating the level of detail of information captured by the optimized embedding \etoken.}  Rows 1-2 show input images and masks. The images are used to optimize \etoken\ via our single-image textual inversion process. Rows 3-5 show examples of 2D images generated using \etoken\ in new prompts, which we hope demonstrate the type of information encoded in \etoken. Rows 6-7 show RealFusion's output, optimized using the prompt ``An image of a \etoken''.
}
\label{fig:method_example}
\end{figure}
\section{Method}\label{s:method}

\newcommand{\bx}{\bm{x}}
\newcommand{\bz}{\bm{z}}
\newcommand{\be}{\bm{e}}
\newcommand{\bzero}{\bm{0}}
\newcommand{\bone}{\bm{I}}
\newcommand{\N}{\mathcal{N}}

We provide an overview and notation for the background material first (\cref{s:background}), and then discuss our \methodname method (\cref{s:core}).

\subsection{Radiance fields and DreamFusion}\label{s:background}

\paragraph{Radiance fields.}

A \emph{radiance field} (RF) is a pair of functions $(\sigma(\bx), c(\bx))$ mapping a 3D point $\bx\in\mathbb{R}^3$ to an opacity value $\sigma(\bx)\in\mathbb{R}_+$ and a color value $c(\bx)\in\mathbb{R}^3$.
The RF is called \emph{neural} when these two functions are implemented by a neural network.

The RF represents the shape and appearance of an object.
In order to generate an image of it, one \emph{renders} the RF using the emission-absorption model.
Let $I\in\mathbb{R}^{3\times H\times W}$ be an image, so that $I(u) \in \mathbb{R}^3$ is the color of pixel $u$.
In order to compute $I(u)$, one casts a ray $r_u$ from the camera center through the pixel, interpreted as a point on the 3D image plane (this implicitly accounts for the camera viewpoint $\pi\in SE(3)$).
Then, one takes a certain number of samples $(\bx_i \in r_u)_{i\in\N},$ for indices $\N=\{1,\dots,N\}$ taken with constant spacing $\Delta$.
The color is obtained as:
\begin{equation}\label{e:ea}
I(u)
= \mathcal{R}(u;\sigma,c)
= \sum_{i\in \N} (T_{i+1} - T_i) c(\bx_i),
~~~
\end{equation}
where
$
T_i = \exp(-\Delta\sum_{j=0}^{i-1} \sigma(\bx_j))
$
is the probability that a photon is transmitted from point $\bx_i$ back to the camera sensor without being absorbed by the material.

Importantly, the rendering function $R(u;\sigma,c)$ is differentiable, which allows training the model by means of a standard optimizer.
Specifically, the RF is fitted to a dataset $\mathcal{D}=\{(I,\pi)\}$ of images $I$ with known camera parameters by minimizing the $L^2$ image reconstruction error
\begin{equation}\label{e:rendloss}
\mathcal{L}_\text{rec}(\sigma,c;\mathcal{D}) =
\frac{1}{|\mathcal{D}|}
\sum_{(I,\pi)\in\mathcal{D}}
\| I - R(\cdot;\sigma,c,\pi) \|^2.
\end{equation}
In order to obtain good quality results, one typically requires a dataset of dozens or hundreds of views.

Here, we consider the case in which we are given \emph{exactly one} input image $I_0$ corresponding to some (unknown) camera $\pi_0$.
In this case, we can also assume \textit{any} standard viewpoint $\pi_0$ for that single camera.
Optimizing \cref{e:rendloss} with a single training image leads to severe over-fitting:
it is straightforward to find a pair $(\sigma,c)$ that has zero loss and yet does not capture any sensible 3D model of the object.
Below we will leverage a pre-trained 2D image prior to (implicitly) dream up novel views of the object and provide the missing information for 3D reconstruction.

\paragraph{Diffusion models.}

A \emph{diffusion model} draws a sample from a probability distribution $p(I)$ by inverting a process that gradually adds noise to the image $I$.
The diffusion process is associated with a variance schedule $\{\beta_t \in (0,1)\}_{t=1}^{T}$, which defines how much noise is added at each time step. The noisy version of sample $I$ at time $t$ can then be written $I_t = \sqrt{\bar{\alpha}_t} I + \sqrt{1-\bar{\alpha}_t} \epsilon$ where
$
\epsilon \sim \mathcal{N}(\bzero,\bone),
$
is a sample from a Gaussian distribution (with the same dimensionality as $I$), $\alpha_t = 1- \beta_t$, and $\bar{\alpha}_t = \prod_{i=1}^t \alpha_i$.
One then learns a denoising neural network
$
\hat \epsilon = \Phi(I_t;t)
$
that takes as input the noisy image $I_t$ and the noise level $t$ and tries to predict the noise component $\epsilon$.

In order to draw a sample from the distribution $p(I)$, one starts by drawing a sample $I_T\sim\mathcal{N}(\bzero,\bone)$.
Then, one progressively denoises the image by iterated application of $\Phi$ according to a specified sampling schedule~\cite{ho2020denoising,song2020denoising,liu2021pseudo}, which terminates with $I_0$ sampled from $p(I)$.

Modern diffusion models are trained on large collections $\mathcal{D'} = \{I\}$ of images by minimizing the loss
\begin{equation}\label{e:diffloss}
\mathcal{L}_\text{diff}(\Phi;\mathcal{D}')
=
\tfrac{1}{|\mathcal{D}'|}
\sum_{I\in\mathcal{D}'}
|| \Phi(\sqrt{\bar{\alpha}_t} I + \sqrt{1-\bar{\alpha}_t} \epsilon, t) - \epsilon ||^2.
\end{equation}

This model can be easily extended to draw samples from a distribution $p(\bx|\be)$ conditioned on a \emph{prompt} $\be$.
Conditioning on the prompt is obtained by adding $\be$ as an additional input of the network $\Phi$, and the strength of conditioning can be controlled via classifier-free guidance~\cite{dhariwal2021diffusion}.

\paragraph{DreamFusion and Score Distillation Sampling (SDS).}

Given a 2D diffusion model $p(I|\be)$ and a prompt $\be$, DreamFusion extracts from it a 3D rendition of the corresponding concept, represented by a RF $(\sigma,c)$.
It does so by randomly sampling a camera parameter $\pi$, rendering a corresponding view $I_\pi$, assessing the likelihood of the view based on the  model $p(I_\pi|\be)$, and updating the RF to increase the likelihood of the generated view based on the model.

In practice, DreamFusion uses the denoiser network as a frozen critic and takes a gradient step
\begin{multline}\label{e:dream}
\nabla_{(\sigma, c)}
\mathcal{L}_\text{SDS}(\sigma,c;\pi,\be,t)
=\\
E_{t,\epsilon}
\Big[
w(t)
(\Phi(\alpha_t I + \sigma_t \epsilon; t,\be) - \epsilon)
\cdot \nabla_{(\sigma, c)} I
\Big],
\end{multline}
where
$
I = R(\cdot;\sigma,c,\pi).
$
is the image rendered from a given viewpoint $\pi$ and prompt $\be$.
This process is called \textit{Score Distillation Sampling} (SDS).

Note that \cref{e:dream} differs from simply optimizing the standard diffusion model objective because it does not include the Jacobian term for $\Phi$. In practice, removing this term both improves generation quality and reduces computational and memory requirements.

One final aspect of \dreamfusion is essential for understanding our contribution in the following section: DreamFusion finds that it is necessary to use classifier-free guidance~\cite{dhariwal2021diffusion} with a very high guidance weight of 100, much larger than one would use for image sampling, in order to obtain good 3D shapes. As a result, the generations tend to have limited diversity; they produce only the most likely objects for a given prompt, which is incompatible with our goal of reconstructing any given object.

\subsection{\Methodname}%
\label{s:core}

\begin{figure*}
\centering
\includegraphics[width=\linewidth]{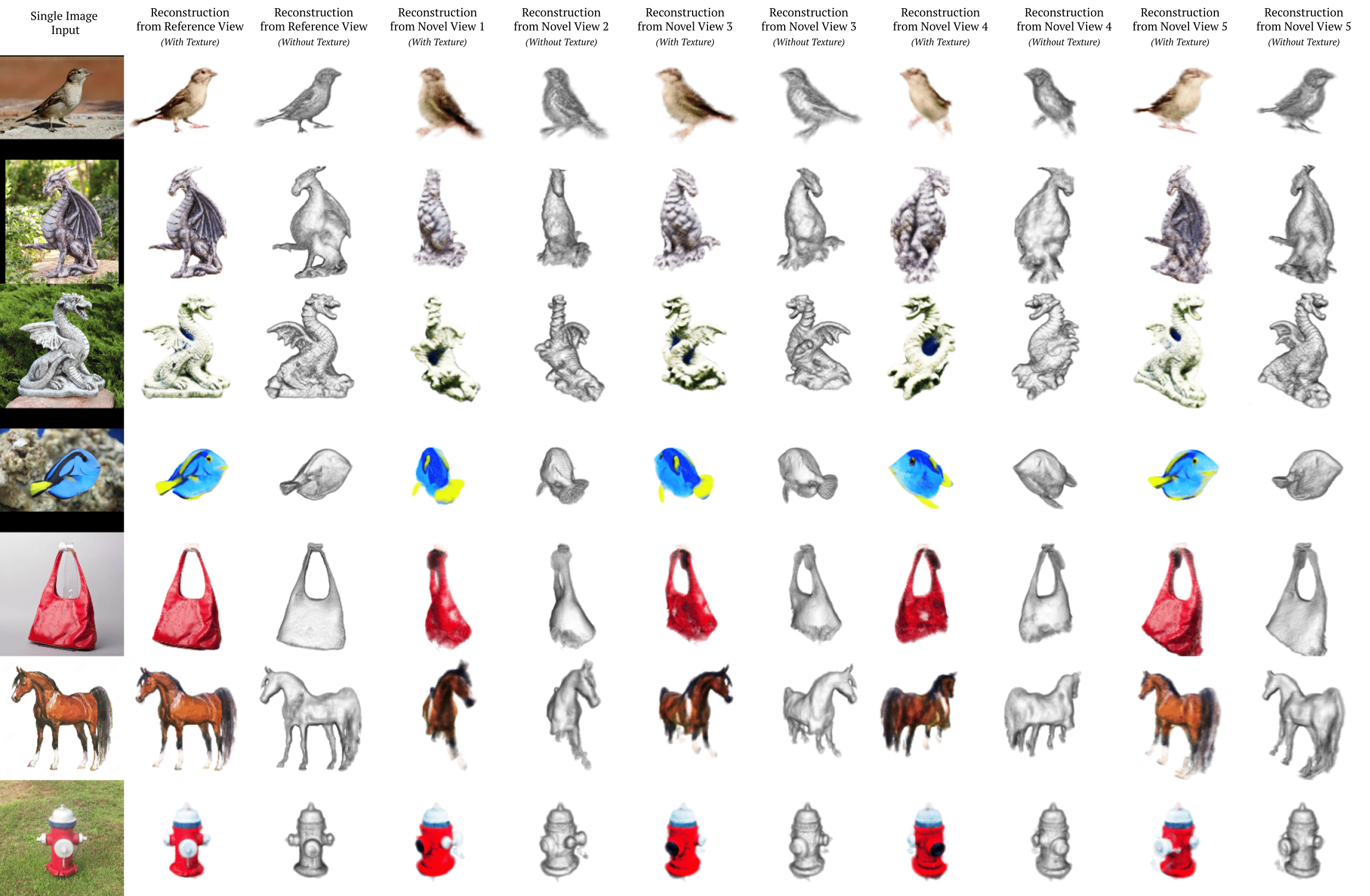}
\captionof{figure}{\textbf{Qualitative results.} \Methodname reconstructions from a single input view.
Each pair of columns shows the textured object and the underlying 3D shape, as a shaded surface.
Different pairs of columns show different viewpoints.}
\label{f:results}
\end{figure*}
\begin{figure}
  \centering
  \includegraphics[width=0.48\textwidth]{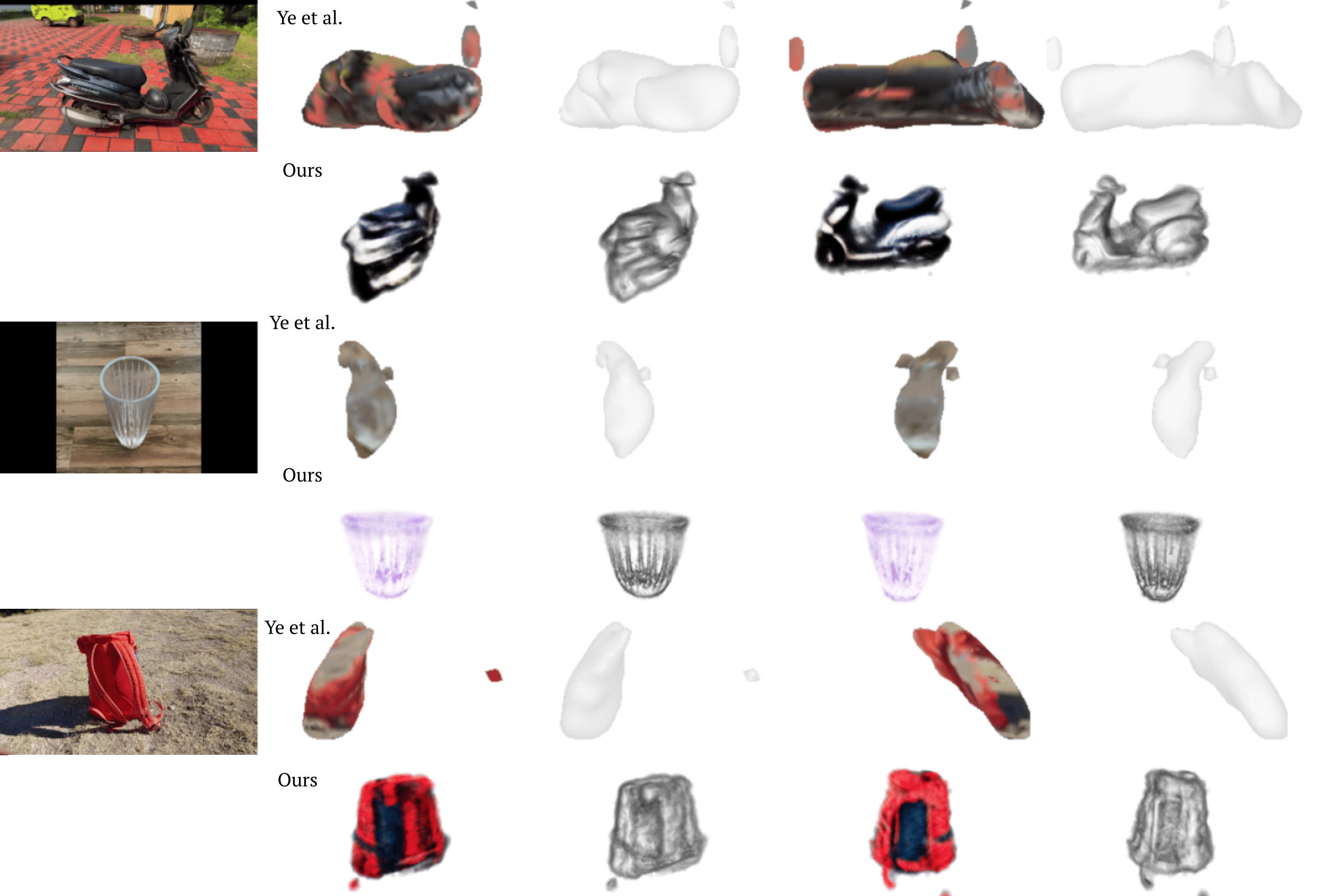}
  \vspace{-2em}
  \caption{
    \textbf{Qualitative comparison with prior work.} We show the results of our method and the category-level method of \cite{ye2021shelf} on real-world images from the CO3D dataset~\cite{reizenstein21co3d}. Each pair of rows show two novel views produced by \cite{ye2021shelf} and our method. For \cite{ye2021shelf}, we use category-specific models for each CO3D category (in this case, motorcycles, cups, and backpacks). Despite not requiring any category-specific information, our method is able to reconstruct objects at a higher level of detail than \cite{ye2021shelf}.
  }
  \label{fig:comparison}
\end{figure}
\begin{figure}
  \centering
  \includegraphics[width=0.47\textwidth]{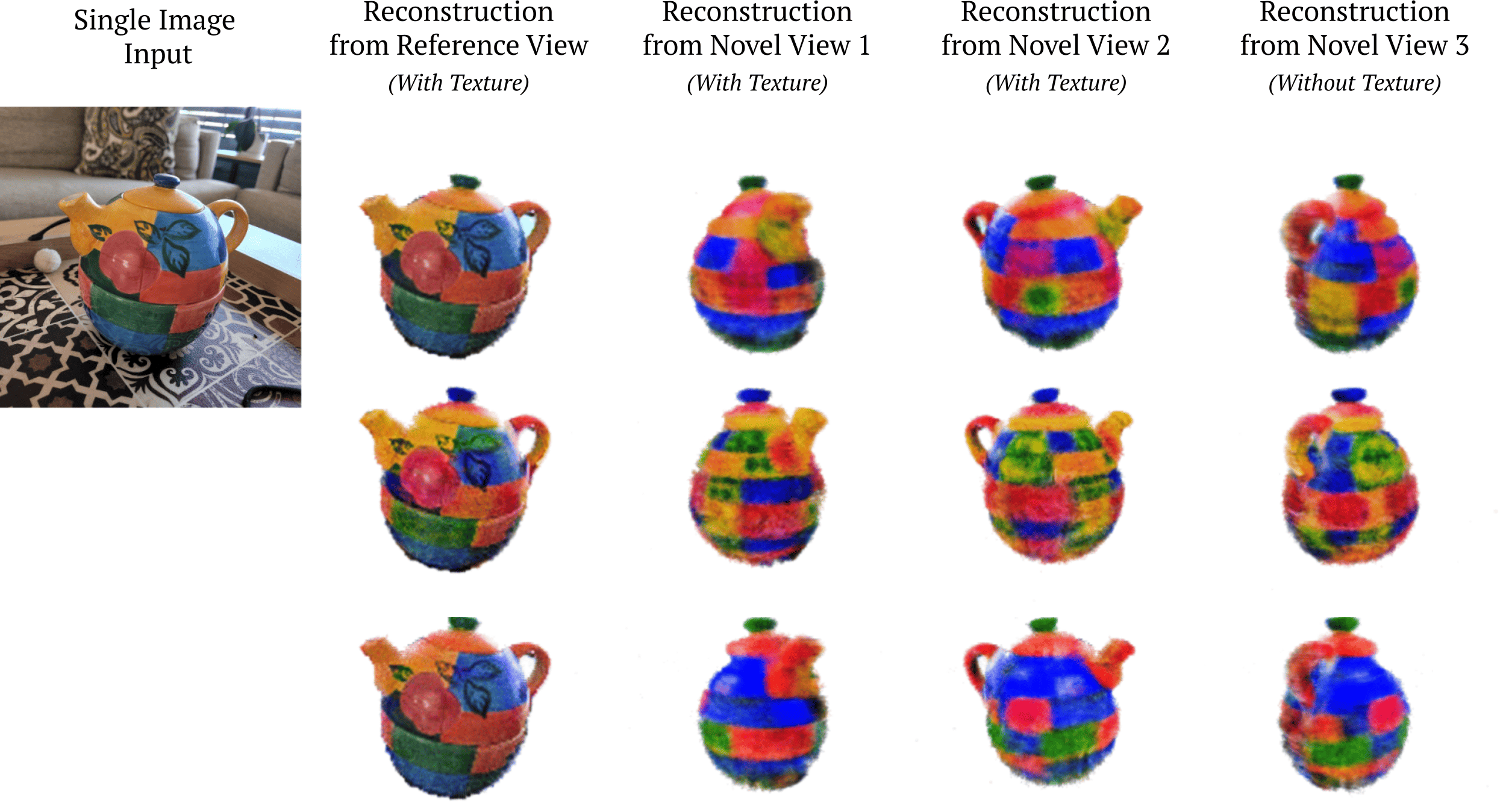}
  \caption{
    \textbf{A demonstration of multi-modal image reconstruction.} Above, we see our method's ability to generate a diverse set of object reconstructions given the same input image.
    In particular, the method produces different textures on the backsides of the generated objects, despite all objects matching the input image from the reference view. 
  }%
  \label{fig:multiple_gens}
  \vspace{-1em}
\end{figure}
\begin{figure*}[t]
\centering
\includegraphics[width=0.95\textwidth]{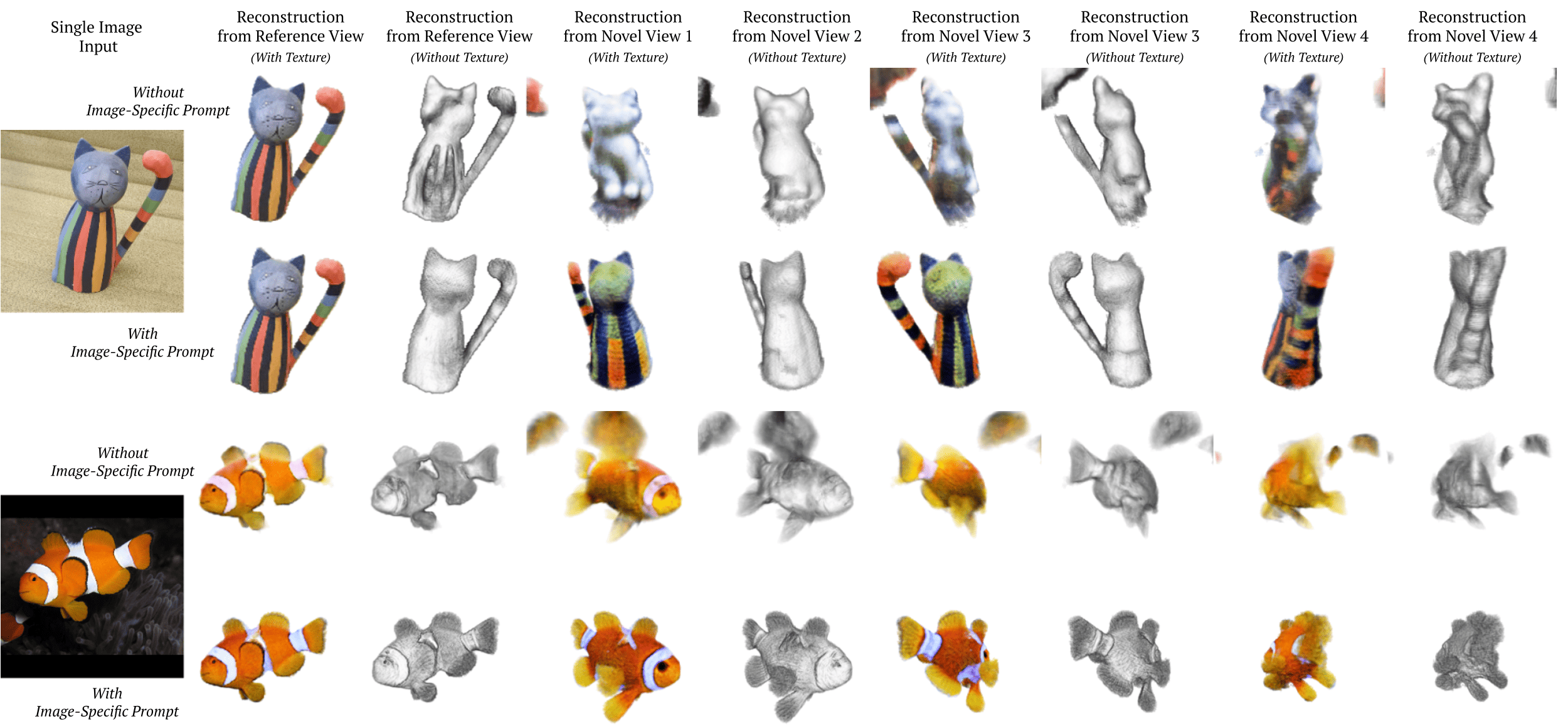}
\caption{
  \textbf{A visualization of the effect of single-image textual inversion on reconstruction quality.} In each pair of rows, the top row shows the result of utilizing a standard text prompt for our diffusion-model-based loss (\eg ``An image of a statue of a cat'').
  The bottom row shows the result of utilizing a text prompt optimized for the input image in a fully-automatic manner; this textual inversion process dramatically improves object reconstruction.
}
\vspace{-1em}
\label{fig:textual_inversion}
\end{figure*}
\begin{figure}
  \centering
  \includegraphics[width=0.47\textwidth]{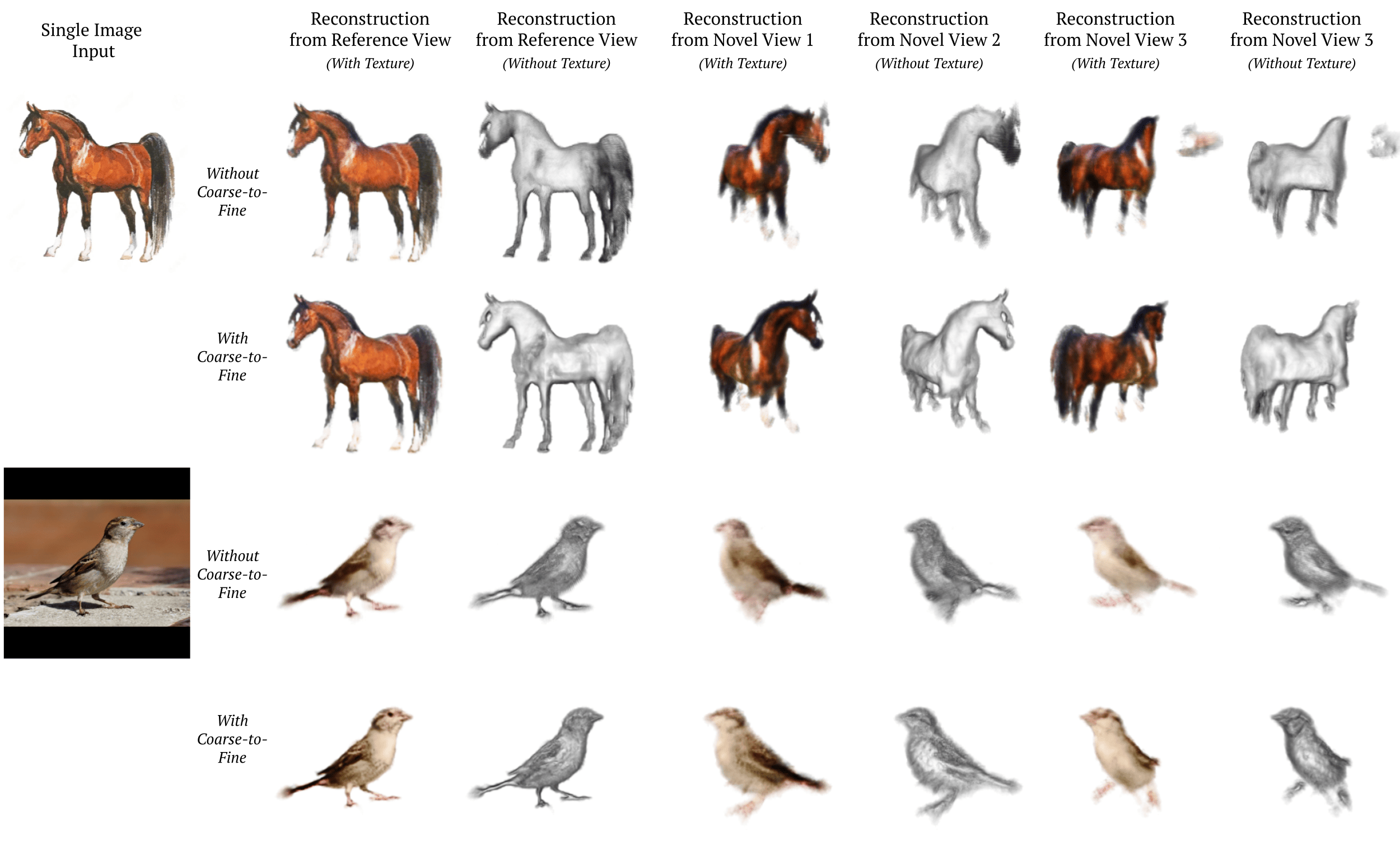}
 \vspace{-1em}
  \caption{
    \textbf{Effect of coarse-to-fine training.}
    The top row of each pair is generated by optimizing all levels of a multi-resolution 3D feature grid from the first optimization step, whereas he bottom row is optimized in a coarse-to-fine manner.
  }
  \label{fig:coarse_to_fine}
\end{figure}
\begin{figure}
  \centering
  \includegraphics[width=0.47\textwidth]{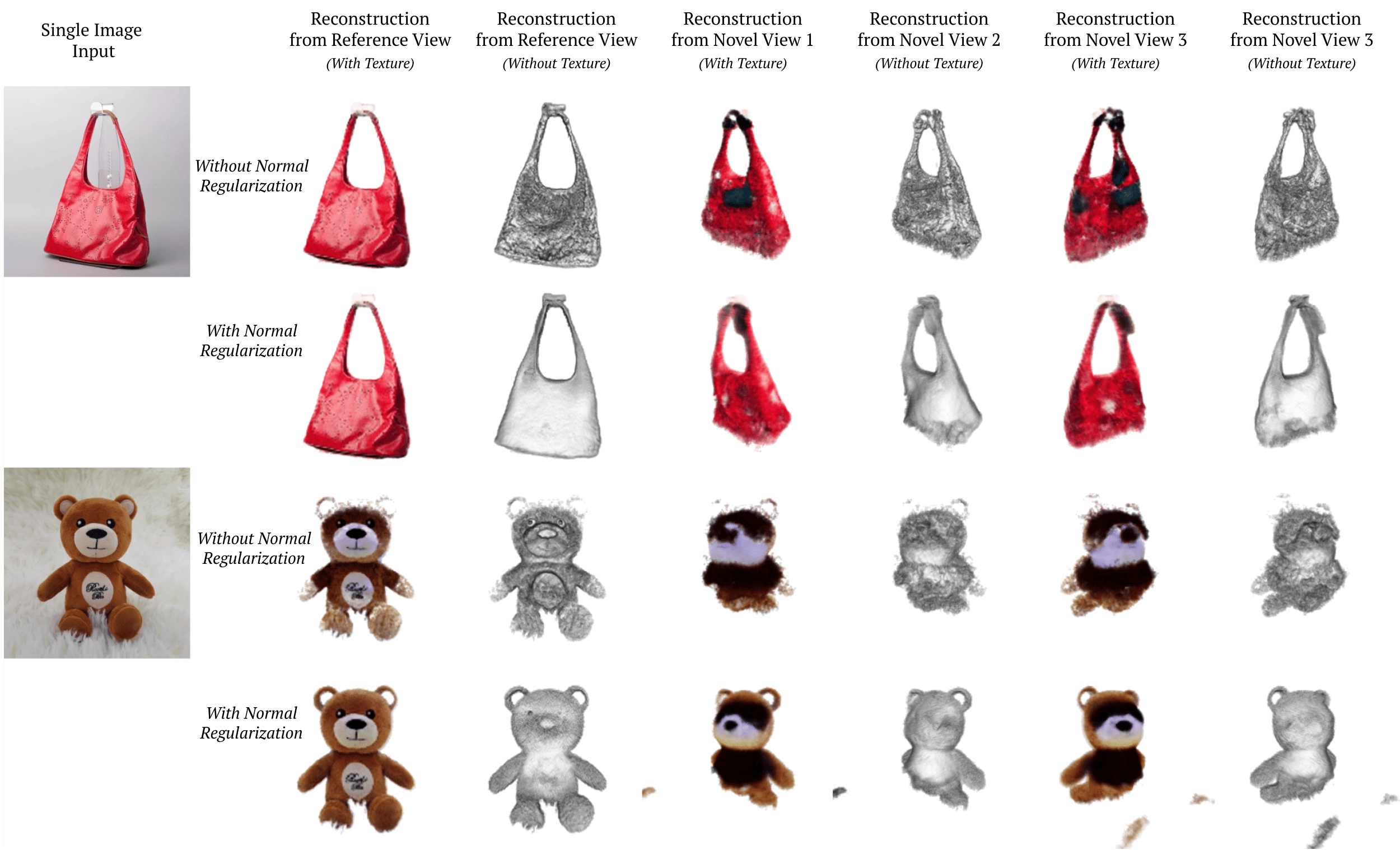}
  \caption{
    \textbf{Effect of normal smoothness on reconstruction quality.}
    Each pair of rows show the reconstruction without and with the normal smoothness regularization term~\eqref{e:normal-smoothness}.
    The regularizer improves the visual appearance of surfaces and reduces the number of irregularities on the surface of reconstructed objects.
    In most cases, we also find that it helps to improve the overall realism of the reconstructed shape. 
  }
  \vspace{-1em}
  \label{fig:normal_smoothness}
\end{figure}
\begin{figure}
  \centering
  \includegraphics[width=0.47\textwidth]{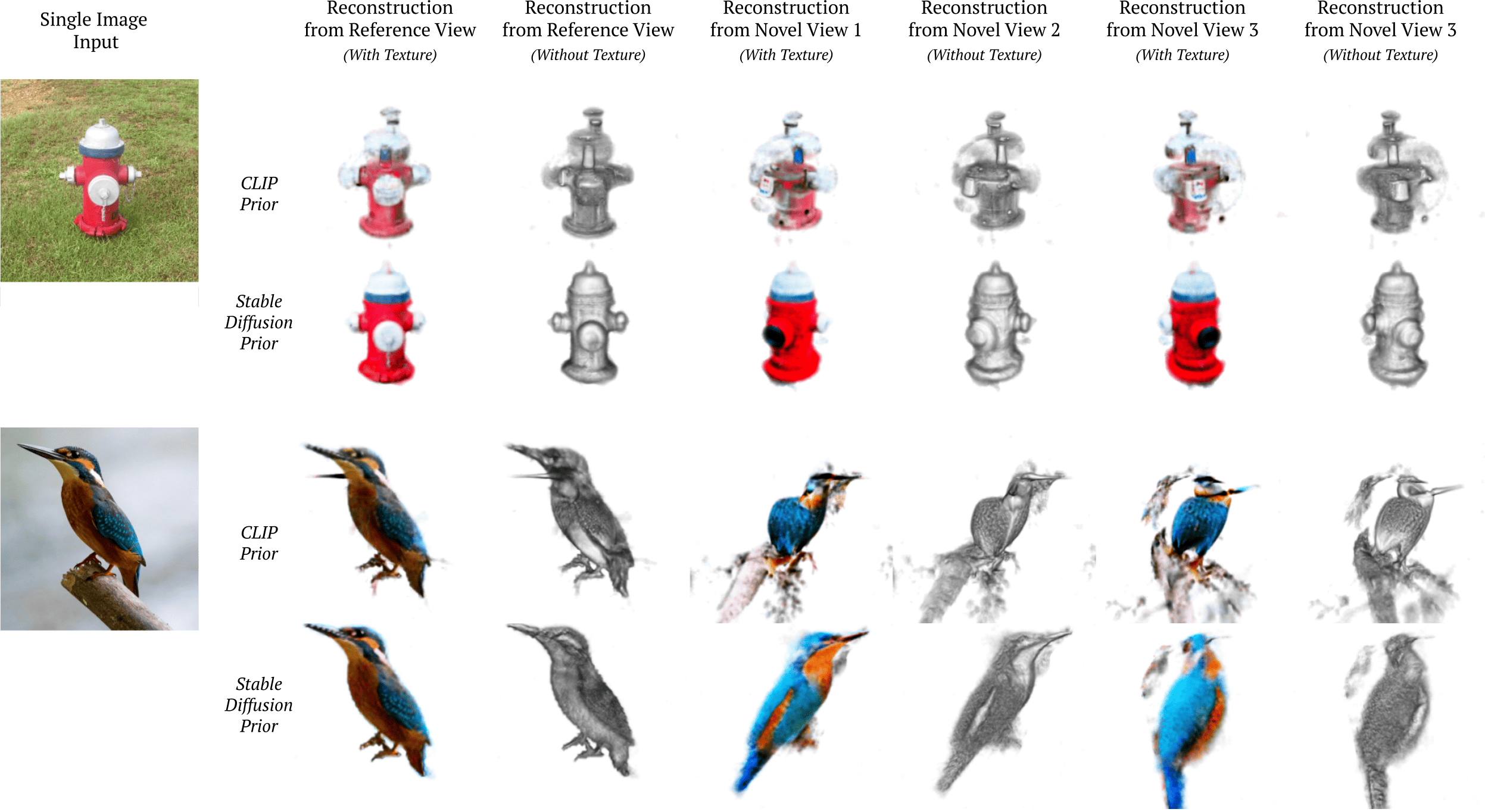}
  \caption{
    \textbf{Comparing Stable Diffusion and CLIP priors.}
    Results from two different priors: Stable Diffusion~\cite{rombach22high-resolution} and CLIP~\cite{radford2021learning}.
    Stable Diffusion yields much higher-quality reconstructions, capturing more plausible object shapes.
  }%
  \label{fig:clip}
\end{figure}
\begin{figure}
  \centering
  \includegraphics[width=0.47\textwidth]{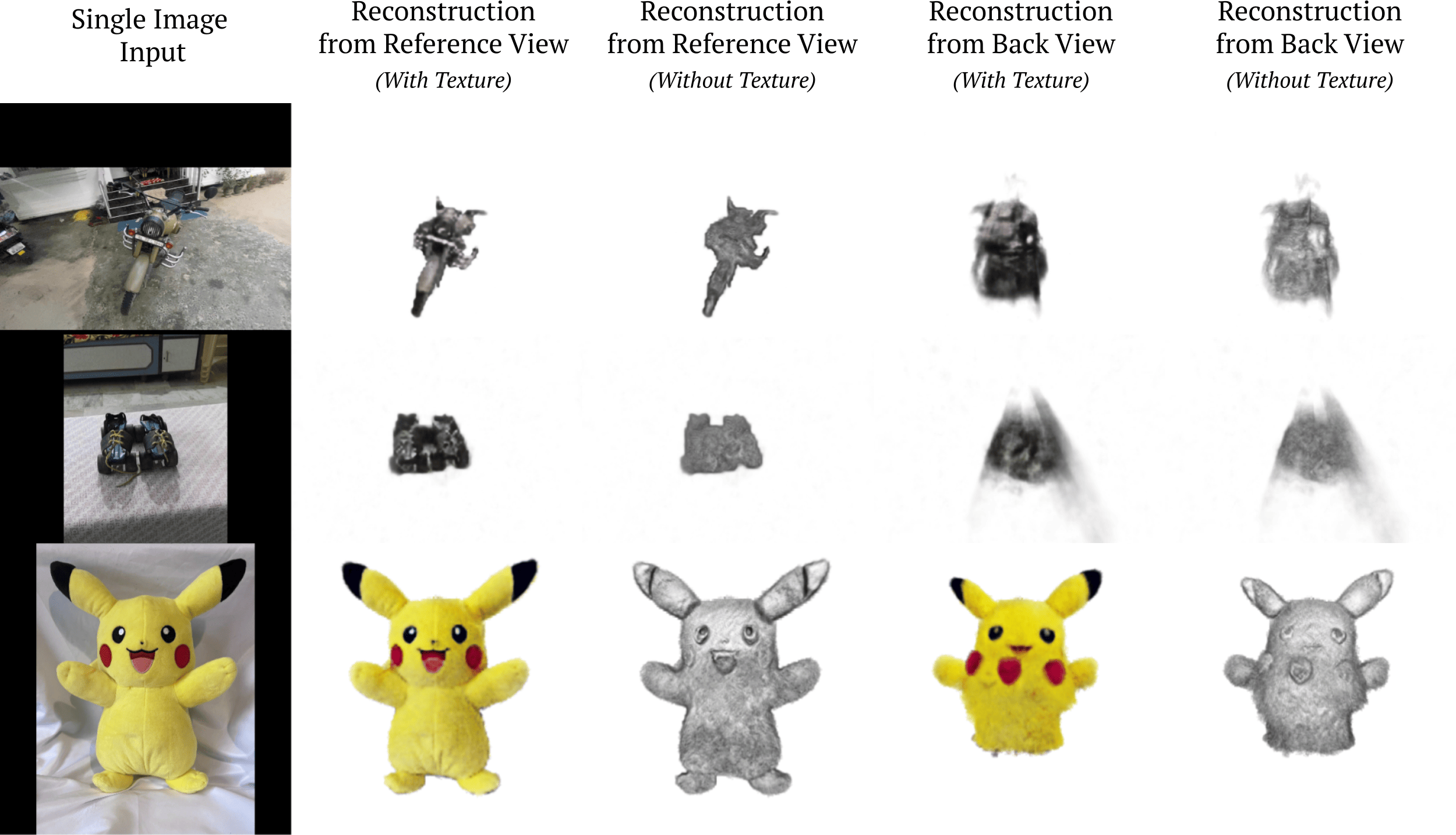}
  \caption{
    \textbf{Failure cases.}
    In the first two examples, the model simply fails to properly reconstruct the object geometry, and produces a semi-transparent scene which lacks a well-defined geometry.
    The third case is different in that the geometry is highly realistic, but the texture paints two Pikachu faces, one on each side of the object; this problem is sometimes called the Janus problem, after the two-faced Roman god.
  }%
  \label{fig:failure_cases}
\end{figure}

Our goal is to reconstruct a 3D model of the object contained in a single image $I_0$, utilizing the prior captured in the diffusion model $\Phi$ to make up for the missing information. We will achieve this by optimizing a radiance field using two simultaneous objectives: (1) a reconstruction objective~\cref{e:rendloss} from a fixed viewpoint, and (2) a SDS-based prior objective~\cref{e:dream} on novel views randomly sampled at each iteration. \Cref{fig:method} provides a diagram of the entire system.

\paragraph{Single-image textual inversion as a substitute for alternative views.}

The most important component of our method
is the use of single-image textual inversion as a substitute for alternative views. Ideally, we would like to condition our reconstruction process on multi-view images of the object in $I_0$, \ie on samples from $p(I|I_0)$. Since these images are not available, we instead synthesize a text prompt $\be^{(I_0)}$ specifically for our image $I_0$ as a proxy for this multi-view information.

Our idea, then, is to engineer a prompt $\be^{(I_0)}$ to provide a useful approximation of $p(I|I_0)$.
We do so by generating random augmentations $g(I_0),$ $g\in G$ of the input image, which serve as pseudo-alternative-views.
We use these augmentations as a mini-dataset
$
\mathcal{D'} = \{ g(I_0) \}_{g\in G}
$
and optimize the diffusion loss~\cref{e:diffloss}
$
\mathcal{L}_\text{diff}(\Phi(\cdot;\be^{(I_0)}))
$
with respect to the prompt $\be^{(I_0)}$, while freezing all other text embeddings and model parameters.

In practice, our prompt is derived automatically from templates like ``an image of a \etoken'', where ``\etoken'' ($=\be^{(I_0)}$) is a new token introduced to the vocabulary of the text encoder of our diffusion model (see \Cref{s:supp_imp_details} for details). Our optimization procedure mirrors and generalizes the recently-proposed textual-inversion method of~\cite{gal2022image}. Differently from~\cite{gal2022image}, we work in the single-image setting and utilize image augmentations for training rather than multiple views.

To help convey the intuition behind \etoken, consider an attempt at reconstructing an image of a fish using the generic text prompt ``An image of a fish'' with losses \cref{e:diffloss,e:dream}. In our experience, this often produces a reconstruction which looks like the input fish from the input viewpoint, but looks like some \textit{different, more-generic} fish from the backside. By contrast, using the prompt ``An image of a \etoken'', the reconstruction resembles the input fish from all angles. An example of exactly this case is shown in \Cref{fig:textual_inversion}.

Finally, \Cref{fig:method_example} demonstrates the amount of detail captured in the embedding \etoken.

\paragraph{Coarse-to-fine training.}

In order to describe our coarse-to-fine training methodology, it is necessary to first briefly introduce our underlying RF model, a InstantNGP~\cite{mueller2022instant}. InstantNGP is a grid-based model which stores features at the vertices of a set of feature grids $\{G_{i}\}_{i=1}^L$ at multiple resolutions. The resolution of these grids is chosen to be a geometric progression between the coarsest and finest resolutions, and feature grids are trained simultaneously.

We choose a InstantNGP over a conventional MLP-based NeRF due to its computational efficiency and training speed. However, the optimization procedure occasionally produces small irregularities on the surface of the object. We find that training in a coarse-to-fine manner helps to alleviate these issues: for the first half of training we only optimize the lower-resolution feature grids $\{G_{i}\}_{i=1}^{L/2}$, and then in the second half of training we optimize all feature grids $\{G_{i}\}_{i=1}^L$. Using this strategy, we obtain the benefits of both efficient training and high-quality results.

\paragraph{Normal vector regularization.}

Next, we introduce a new regularization term to encourage our geometry to have smooth normals. 
The introduction of this term is motivated by the observation that our RF model occasionally generated noisy-looking surfaces with low-level artifacts.
To address these artifacts, we encourage our RF to have smoothly varying normal vectors. Notably, we perform this regularization in \textit{2D} rather than in 3D.

At each iteration, in addition to computing RGB and opacity values, we also compute normals for each point along the ray and aggregate these via the raymarching equation to obtain normals $N \in \mathcal{R}^{H\times W \times 3}$.\footnote{Normals may be computed either by taking the gradient of the density field or by using finite differences. We found that using finite differences worked well in practice.}
Our loss is:
\begin{equation}\label{e:normal-smoothness}
\mathcal{L}_\mathrm{normals} = \| N - \operatorname{stopgrad}(\text{blur}(N, k)) \|^2
\end{equation}
where $\text{stopgrad}$ is a stop-gradient operation and $\text{blur}(\cdot, k)$ is a Gaussian blur with kernel size $k$ (we use $k=9$).

Although it may be more common to regularize normals in 3D, we found that operating in 2D reduced the variance of the regularization term and led to superior results.

\paragraph{Mask loss.}

In addition to the input image, our model also utilizes a mask of the object that one wishes to reconstruct. In practice, we use an off-the-shelf image matting model to obtain this mask for all images.

We incorporate this mask in a simple manner by adding a simple $L^2$ loss term on the difference between the rendered opacities from the fixed reference viewpoint  $\mathcal{R}(\sigma, \pi_0) \in \mathcal{R}^{H\times W}$ and the object mask $M$:
$\mathcal{L}_\text{rec,mask} = ||O - M||^2$
Our final objective then consists of four terms:
\begin{multline}
\nabla_{\sigma,c} \mathcal{L} =
\nabla \mathcal{L}_\text{SDS} +
\lambda_{\text{normals}} \cdot \nabla \mathcal{L}_{\text{normals}} \\ +
\lambda_{\text{image}} \cdot \nabla \mathcal{L}_{\text{image}} +
\lambda_{\text{mask}} \cdot \nabla \mathcal{L}_{\text{mask}}
\end{multline}
where the top line in the equation above corresponds to our prior objective and the bottom line corresponds to our reconstruction objective.

\section{Experiments}\label{s:experiments}

\subsection{Implementation details}

Regarding hyperparameters, we use \textit{essentially the same set of hyper-parameters for all experiments}---there is no per-scene hyper-parameter optimization.\footnote{There is one small exception to this rule, which is that for a few number of images where the camera angle was clearly at an angle higher than 15$^\circ$, we took a camera angle of 30 or 40$\deg$.}.
For our diffusion model prior, we employ the open-source \emph{Stable Diffusion} model~\cite{rombach22high-resolution} trained on the LAION~\cite{schuhmann2022laion} dataset of text-image pairs.
For our InstantNGP~\cite{mueller2022instant} model, we use a model with 16 resolution levels, a feature dimension of 2, and a maximum resolution of 2048, trained in a coarse-to-fine manner as explained above.

The camera for reconstruction is placed looking at the origin on a sphere of radius $1.8$, at an angle of 15$\deg$ above the plane.
At each optimization step, we first render from the reconstruction camera and compute our reconstruction losses $\mathcal{L}_\mathrm{rec}$ and $\mathcal{L}_{\mathrm{rec},\mathrm{mask}}$.
We then render from a randomly sampled camera to obtain a novel view, and use this view for $\mathcal{L}_\mathrm{sds}$ and $\mathcal{L}_\mathrm{normals}$. We use $\lambda_{\text{image}} = 5.0$, $\lambda_{\text{mask}} = 0.5$, and  $\lambda_{\text{normal}} = 0.5$.

Regarding camera sampling, lighting, and shading, we keep nearly all parameters the same as~\cite{poole22dreamfusion:}. This includes the use of diffuse and textureless shading stochastic throughout the course of optimization, after an initial warmup period of albedo-only shading. Complete details regarding this and other aspects of our training setup are provided in the supplementary material.

\begin{table}[t]
\centering
\small
\caption{\textbf{Quantitative comparison.} We compare our method with Shelf-Supervised~\cite{ye2021shelf} on seven object categories. 
The F-score and CLIP-similarity metrics are designed to measure the quality of reconstruction shape and appearance, respectively. For both metrics, higher is better. Metrics are averaged over three images per category. Our method outperforms \cite{ye2021shelf} in aggregate, despite the fact that 
\cite{ye2021shelf} uses a \textit{different category-specific model} for each category.
}%
\label{table:comparison_both}
\begin{tabular}{l|cc|cc}
\toprule
           & \multicolumn{2}{c}{\textit{F-score}} & \multicolumn{2}{|c}{\textit{CLIP-similarity}} \\
\midrule
           & {}\cite{ye2021shelf}                 & Ours                                            & {}\cite{ye2021shelf} & Ours \\
\midrule
Backpack   & 7.58                                 & \bfseries 12.22                                 & 0.72                 & \bfseries 0.74 \\
Chair      & 8.26                                 & \bfseries 10.23                                 & 0.65                 & \bfseries 0.76 \\
Motorcycle & 8.66                                 & \bfseries 8.72                                  & 0.69                 & \bfseries 0.70 \\
Orange     & 6.27                                 & \bfseries 10.16                                 & 0.71                 & \bfseries 0.74 \\
Skateboard & \bfseries 7.74                       & 5.89                                            & \bfseries 0.74       & \bfseries 0.74 \\
Teddybear  & \bfseries 12.89                      & 10.08                                           & 0.73                 & \bfseries 0.82 \\
Vase       & 6.30                                 & \bfseries 9.72                                  & 0.69                 & \bfseries 0.71 \\
\midrule
Mean       & 8.24                                 & \bfseries 9.58                                  & 0.70                 & \bfseries 0.74 \\
\bottomrule
\end{tabular}
\end{table}

\subsection{Quantitative results}

There are only few methods that attempt to reconstruct arbitrary objects in 3D.
The most recent and best-performing of these is Shelf-Supervised Mesh Prediction~\cite{ye21shelf-supervised}, which we compare here.
They provide 50 pretrained category-level models for 50 different categories in OpenImages~\cite{OpenImages}.
Since we aim to compute metrics using 3D or multi-view ground truth, we evaluate on seven categories in the CO3D dataset~\cite{reizenstein21common}
with corresponding OpenImages categories.
For each of these seven categories, we select three images at random and run both \methodname and Shelf-Supervised to obtain reconstructions.

We first test the quality of the recovered 3D shape in \cref{fig:comparison}.
Shelf-Supervised directly predicts a mesh.
We extract one from our predicted radiance fields using marching cubes.
CO3D comes with sparse point-cloud reconstruction of the objects obtained using multi-view geometry.
For evaluation, we sample points from the reconstructed meshes and align them optimally with the ground truth point cloud by first estimating a scaling factor and then using Iterated Closest Point (ICP). Finally, we compute F-score with threshold $0.05$ to measure the distance between the predicted and ground truth point clouds. Results are shown in \cref{table:comparison_both}.

In order to evaluate the quality of the reproduced appearance, we also compare novel-view renderings from our and their method (\cref{table:comparison_both}).
Ideally, these renderings should produce views that are visually close to the real views.
In order to test this hypothesis, we check whether the generated views are close or not to the other views given in CO3D.
We then report the CLIP embedding similarity of the generated images with respect to the closest CO3D view available (\ie the view with maximum similarity).

\subsection{Qualitative results}
\Cref{f:results} shows additional qualitative results from multiple viewpoints.
Having a single image of an object means that several 3D reconstructions are possible.
\Cref{fig:multiple_gens} explores the ability of \methodname to sample the space of possible solutions by repeating the reconstruction several times, starting from the same input image.
There is little variance in the reconstructions of the front of the object, but quite a large variance for its back, as expected.

\Cref{fig:failure_cases} shows two typical failure modes of \methodname: in some cases the model fails to converge, and in others it copies the front view to the back of the object, even if this is not semantically correct.

\subsection{Analysis and Ablations}

One of the key components of \methodname is our use of single-image textual inversion, which allows the model to correctly imagine novel views of a specific object.
\Cref{fig:textual_inversion} shows that this component plays indeed a critical role in the quality of the reconstructions.
Without texual inversion, the model often reconstructs the backside of the object in the form of a generic instance from the object category.
For example, the backside of the cat statue in the top row of \cref{fig:textual_inversion} is essentially a different statue of a more generic-looking cat, whereas the model trained with textual inversion resembles the true object from all angles.

Other components of the model are also significant.
\Cref{fig:normal_smoothness} shows that the normal smoothness regularizer of~\cref{e:normal-smoothness} results in smoother, more realistic meshes and reduces the number of artifacts.
\Cref{fig:coarse_to_fine} shows that coarse-to-fine optimization reduces the presence of low-level artifacts and results in smoother, visually pleasing surfaces.
\cref{fig:clip} shows that using Stable Diffusion works significantly better than relying on an alternative such as CLIP. %

\section{Conclusions}\label{s:conclusions}

We have introduced \methodname, a new approach to obtain full 360$^\circ$ photographic reconstructions of any object given a single image of it.
Given an off-the-shelf diffusion model trained using only 2D images and no special supervision for 3D reconstruction, as well as a single view of the target object, we have shown how to select the model prompt to imagine other views of the object.
We have used this conditional prior to learn an efficient, multi-scale radiance field representation of the reconstructed object, incorporating an additional regularizer to smooth out the reconstructed surface.
The resulting method can generate plausible 3D reconstructions of objects captured in the wild which are faithful to the input image.
Future works include specializing the diffusion model for the task of new-view synthesis and incorporating dynamics to reconstruct animated 3D scenes.

\paragraph{Ethics.}

We use the CO3D dataset in a manner compatible with their terms.
CO3D does not contain personal information.
Please see \url{https://www.robots.ox.ac.uk/~vedaldi/research/union/ethics.html} for further information on ethics.

\paragraph{Acknowledgments.}

L\@. M\@. K\@. is supported by the Rhodes Trust.
A\@. V\@., I\@. L\@. and C.R. are supported by ERC-UNION-CoG-101001212.
C\@. R\@. is also supported by VisualAI EP/T028572/1.

{\small\bibliographystyle{ieee_fullname}\bibliography{refs,refs_diffusion,refs_shelf,vedaldi_general,vedaldi_specific}}

\clearpage

\appendix
\section{Implementation Details}\label{s:supp_imp_details}

In this section, we provide full implementation details which were omitted from the main text due to space constraints. Most of these details follow \cite{poole22dreamfusion:}, but a few are slightly modified.

\paragraph{Shading.} We consider three different types of shading: albedo, diffuse, and textureless. For albedo, we simply render the RGB color of each ray as given by our model:
\[ I(u) = I_{\rho}(u) = \mathcal{R}(u; \sigma, c) \]
For diffuse, we also compute the surface normal $n$ as the normalized negative gradient of the density with respect to $u$. Then, given a point light $l$ with color $l_{\rho}$ and an ambient light with color $l_{a}$, we render
\[
I(u) = I_{\rho}(u) \circ (l_{\rho} \circ \max(0, n \cdot \tfrac{l - u}{||l-u||} + l_{a}))
\]
For textureless, we use the same equation with $I_{\rho}(u)$ replaced by white $(1,1,1)$.

For the reconstruction view, we only use albedo shading. For the random view (i.e. the view used for the prior objectives), we use albedo shading for the first 1000 steps of training by setting $l_{a} = 1.0$ and $l_{\rho} = 0.0$. Afterwards we use $l_{a} = 0.1$ and $l_{\rho} = 0.9$, and we select stochastically between albedo, diffuse, and textureless with probabilities $0.2$, $0.4$, and $0.4$, respectively. 

We obtain the surface normal using finite differences: 
\[
n = \frac{1}{2 \cdot \eps} \begin{pmatrix}
I(u + \eps_x) - I(u - \eps_x) \\
I(u + \eps_y) - I(u - \eps_y) \\
I(u + \eps_z) - I(u - \eps_z)
\end{pmatrix}
\]
where 
$\eps_x = (\eps, 0, 0)$, 
$\eps_y = (0, \eps, 0)$, and 
$\eps_z = (0, 0, \eps)$

\paragraph{Density bias.} As in \cite{poole22dreamfusion:}, we add a small Gaussian blob of density to the origin of the scene in order to assist with the early stages of optimization. This density takes the form 
\[
\sigma_{\text{init}}(\mu) = \lambda \cdot e^{-||\mu||^2 / (2 \nu^2)}
\]
with $\lambda = 5$ and $\nu = 0.2$.

\paragraph{Camera.} 

The fixed camera for reconstruction is placed at a distance of $1.8$ from the origin, oriented toward the origin, at an elevation of $15\degrees$ above the horizontal plane. For a small number of scenes in which the object of interest is clearly seen from overhead, the reconstruction camera is placed at an elevation of $40\degrees$.

The camera for the prior objectives is sampled randomly at each iteration. Its distance from the origin is sampled uniformly from $[1.0, 1.5]$. Its azimuthal angle is sampled uniformly at random from the $360^{\circ}$ around the object. Its elevation is sampled uniformly in degree space from $-10\degrees$ to $90\degrees$ with probability $0.5$ and uniformly on the upper hemisphere with probability $0.5$. The field of view is uniformly sampled between $40$ and $70$. The camera is oriented toward the origin. Additionally, every tenth iteration, we place the prior camera near the reconstruction camera: its location is sampled from the prior camera's location perturbed by Gaussian noise with mean $0$ and variance $1$.

\paragraph{Lighting.} 

We sample the position of the point light by adding a noise vector $\eta\sim\mathcal{N}(0,1)$ to the position of the prior camera. 

\paragraph{View-Dependent Prompt.} 

We add a view-dependent suffix to our text prompt based on the location of the prior camera relative to the reconstruction camera. If the prior camera is placed at an elevation of above $60\degrees$, the text prompt receives the suffix ``overhead view.'' If it is at an elevation below $0\degrees$, the text receives ``bottom view.'' Otherwise, for azimuthal angles of $\pm 30\degrees$, $\pm 30-90\degrees$, or $\pm 90-180\degrees$ in either direction of the reconstruction camera, it receives the suffices ``front view,'' ``side view,'' or ``bottom view,'' respectively.

\begin{figure*}[t]
\centering
\small
\begin{lstlisting}[language=python]
    transform = T.Compose([
        T.RandomApply([T.RandomRotation(degrees=10, fill=255)], p=0.75),
        T.RandomResizedCrop(image_size, scale=(0.70, 1.3)),
        T.RandomApply([T.ColorJitter(0.04, 0.04, 0.04, 0.04)], p=0.75),
        T.RandomGrayscale(p=0.10),
        T.RandomApply([T.GaussianBlur(5, (0.1, 2))], p=0.10),
        T.RandomHorizontalFlip(),
    ])
\end{lstlisting}
\caption{PyTorch code for the image augmentations used for single-image textual inversion.}
\label{fig:augs}
\end{figure*}

\paragraph{InstantNGP.} 

Our InstantNGP parameterizes the density and albedo inside a bounding box around the origin with side length $0.75$. It is a multi-resolution feature grid with 16 levels. With coarse-to-fine training, only the first 8 (lowest-resolution) levels are used during the first half of training, while the others are masked with zeros. Each feature grid has dimensionality $2$. The features from these grids are stacked and fed to a 3-layer MLP with $64$ hidden units. 

\paragraph{Rendering and diffusion prior.} We render at resolution $96$px. Since Stable Diffusion~\cite{rombach22high-resolution} is designed for images with resolution $512$px, we upsample renders to $512$px before passing them to the Stable Diffusion latent space encoder (i.e. the VAE). We add noise in latent space, sampling $t \sim \mathcal{U}(0.02, 0.98)$. We use classifier-free guidance strength $100$. We found that results with classifier-free guidance strength above $30$ produced good results; below $30$ led to many more geometric deformities. Although we do not backpropagate through the Stable Diffusion UNet for $\mathcal{L}_{\text{SDS}}$, we do backpropagate through the latent space encoder. 

\paragraph{Optimization.} We optimize using the Adam~\cite{kingma14adam:} optimizer with learning rate $1e-3$ for $5000$ iterations. The optimization process takes approximately $45$ minutes on a single V100 GPU.

\paragraph{Background model.} For our background model, we use a two-layer MLP which takes the viewing direction as input. This model is purposefully weak, such that the model cannot trivially optimize its objectives by using the background.

\paragraph{Additional regularizers.}

We additionally employ two regularizers on our density field. The first is the orientation loss from Ref-NeRF~\cite{verbin2022ref}, also used in DreamFusion~\cite{poole22dreamfusion:}, for which we use $\lambda_{\text{orient}} = 0.01$. The second is an entropy loss which encourages points to be either fully transparent or fully opaque: $\mathcal{L}_{\text{entropy}} = (w \cdot \log_2 (w) - (1-w) \cdot \log_2 (1-w)$
where $w$ is the cumulative sum of density weights computed as part of the NeRF rendering equation (Equation 1).

\paragraph{Single-image textual inversion.}

Our single-image textual inversion step, which is a variant of textual inversion~\cite{gal2022image}, entails optimizing a token $\textbf{e}$ introduced into the diffusion model text encode to match an input image. The key to making this optimization successful given only a single image is the use of heavy image augmentations, shown in ~\cref{fig:augs}. We optimize using these augmentations for a total of 3000 steps using the Adam optimizer~\cite{kingma14adam:} with image size $512$px, batch size 16, learning rate $5\cdot 10^{-4}$, and weight decay $1\cdot 10^{-2}$.

The embedding $\textbf{e}$ can be initialized either randomly, manually (by selecting a token from the vocabulary that matches the object), or using an automated method. 

One automated method that we found to be successful was to use CLIP (which is also the text encoder of the Stable Diffusion model) to infer a starting token to initialize the inversion procedure. For this automated procedure, we begin by considering the set of all tokens in the CLIP text tokenizer which are nouns, according to the WordNet~\cite{wordnet} database. We use only nouns because we aim to reconstruct objects, not reproduce styles or visual properties. We then compute text embeddings for captions of the form ``An image of a $\langle\texttt{token}\rangle$" using each of these tokens. Separately, we compute the image embedding for the input image. Finally, we take the token whose caption is most similar to the image embedding as initialization for our textual inversion procedure. 

We use the manual initialization method for the examples in the main paper and we use the automated initialization method for the examples in the supplemental material (i.e. those included below).

\section{Method diagram}

We provide a diagram illustrating our method in \cref{fig:method}.

\section{Additional Qualitative Examples}

\begin{figure*}[t]
\centering
\includegraphics[width=0.98\textwidth]{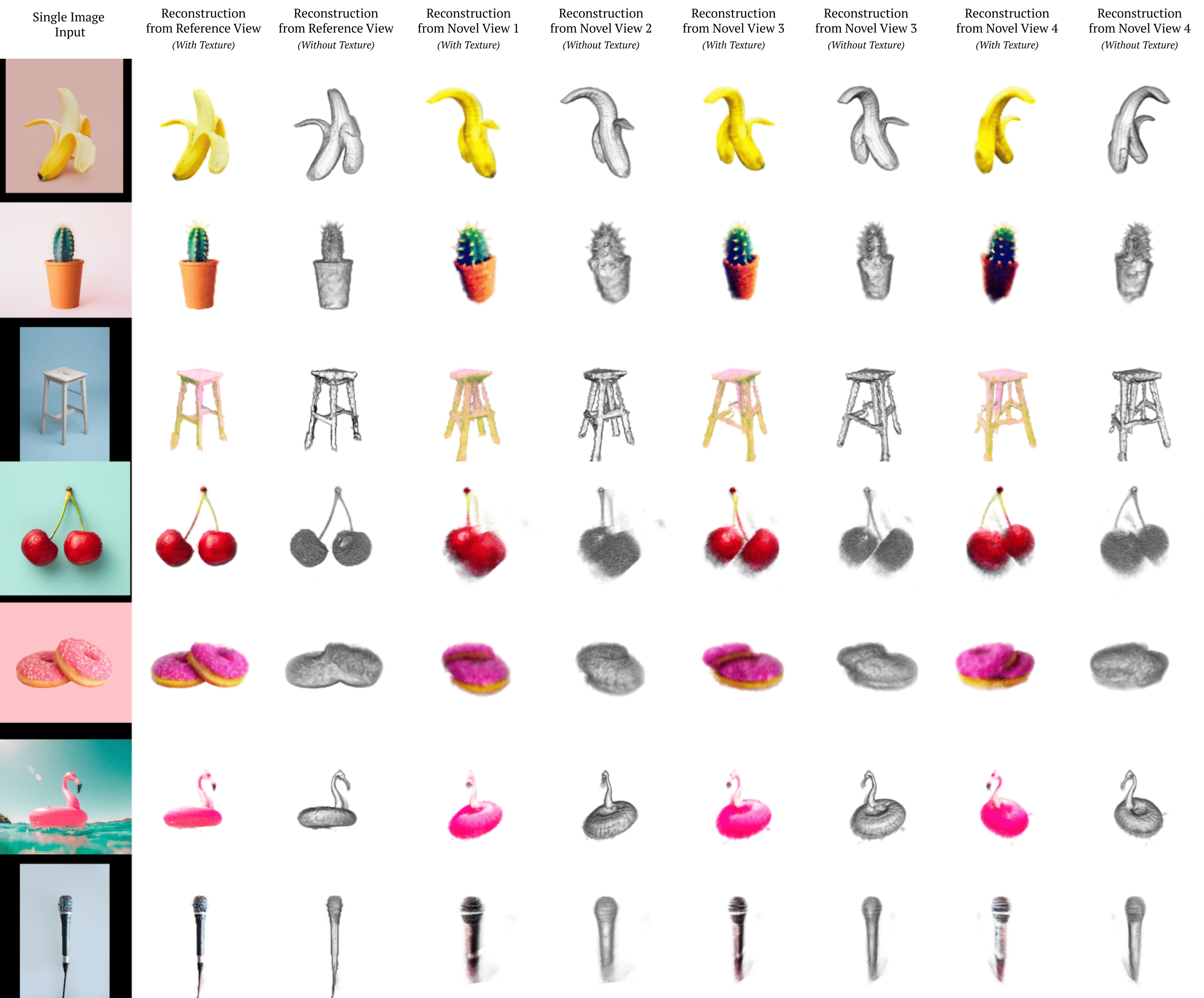}
\caption{
    \textbf{Additional qualitative examples.} This figure presents additional qualitative examples from our model. The first column shows the input image. The second column shows the reconstruction from the reference viewpoint. The following columns show renders from novel viewpoints, demonstrating that our model is able to reconstruct plausible object shapes. 
}
\label{fig:supplementary_results_1}
\end{figure*}

In \cref{fig:supplementary_results_1}, we show additional examples of reconstructions from our model. We see that our method is often able to reconstruct plausible geometries and object backsides. 

\section{Additional Comparisons}

We provide additional comparisons to recent single-view reconstruction methods on the lego scene from the synthetic NeRF~\cite{mildenhall2020nerf} dataset. We compare on the special test set created by SinNeRF~\cite{Xu_2022_SinNeRF}, which consists of 60 views very close to the reference view. We emphasize that our method is not tailored to this setting, whereas the other methods are designed specifically for it. For example, some other methods work by warping the input image, which only performs well for novel views close to the reference view. 

\begin{table}[h!]
\centering
\small
\caption{\textbf{Novel view synthesis comparison.} A comparison of RealFusion against recent single-view reconstruction methods on the task of novel view synthesis on the synthetic lego scene from NeRF~\cite{mildenhall2020nerf}. These numbers are computed on the test set rendered by SinNeRF~\cite{Xu_2022_SinNeRF}, which contains 60 views very close to the reference view. This is a setting highly favorable to methods that use depth supervision, such as DS-NeRF~ and SinNeRF~.
}%
\begin{tabular}{l|c|ccc}
                                          & \textit{Depth?} & \textit{PSNR} & \textit{SSIM} & \textit{LPIPS} \\ \midrule
\textit{PixelNeRF~\cite{yu2021pixelnerf}} &                & 14.3          & 0.72          & 0.22           \\
\textit{DietNeRF~\cite{Jain_2021_ICCV}}   &                & 15.0          & 0.72          & 0.20           \\
\textit{DS-NeRF~\cite{kangle2021dsnerf}}  & \cmark         & 16.6          & 0.77          & 0.16           \\
\textit{SinNeRF~\cite{Xu_2022_SinNeRF}}   & \cmark         & \textbf{21.0} & \textbf{0.82} & \textbf{0.09}  \\
\textit{RealFusion}                       &                & 16.5          & 0.76          & 0.25          
\end{tabular}
\label{table:comparison_synthetic}

\end{table}
    
\section{Text-to-Image-to-3D}

In this section, we explore the idea of reconstructing a 3D object from a text prompt alone by first using the text prompt to generate an image, and then reconstructing this image using \methodname. 

We show examples of text-to-image-to-3D generation in \cref{fig:supplementary_results_2_text2image}.

Compared to the one-step procedure of \cite{poole22dreamfusion:} (i.e. text-to-3D), this two-step procedure (i.e. text-to-image-to-3D) has the advantage that it may be easier for users to control. Under our setup, users can first sample a large number of images from a 2D diffusion model such as Stable Diffusion, select their desired image, and then lift it to 3D using \methodname. It is possible that this setup could help help address the issue of diversity of generation discussed in ~\cite{poole22dreamfusion:}. Additionally, tn this setting, we find that it is usually not necessary to use single-image textual inversion, since the images sampled in the first stage are already extremely well-aligned with their respective prompts. 

\section{Analysis of Failure Cases}

In \cref{fig:supplementary_results_3_failure}, we show additional examples of failure cases from our model. Below, we analyzed what we find to be our three most common failure cases. The techniques we apply in \Methodname (single-image textual inversion, normals smoothing, and coarse-to-fine training) make these failure cases less frequent and less severe, but they still occur on various images.

\paragraph{Neural fields lacking well-defined geometry.}

One failure case of our method consists of the generation of a semi-transparent neural field which does not have a well-defined geometry. These fields tend to look like the input image when seen from the reference viewpoint, but do not resemble plausible objects when seen from other viewpoints. We note that this behavior is extremely common when using CLIP as a prior model, but it occurs occasionally even when using Stable Diffusion and $\mathcal{L}_{\text{SDS}}$.

\paragraph{Floaters.} \label{s:floaters}

Another failure case involves ``floaters,'' or disconnected parts of the scene which appear close to the camera. These floaters sometimes appear in front of the reference view as to make the corresponding render look like the input image. Without image-specific prompts, these floaters are a very big issue, appearing in the majority of reconstructions. When using image-specific prompts, the issue of floaters is greatly (but not entirely) alleviated. 

\paragraph{The Janus Problem.}

Named after the two-faced Roman god Janus, the ``Janus problem'' refers to reconstructions which have two or more faces. This problem arises because the loss function tries to make the render of every view look like the input image, at least to a certain extent. 

\begin{figure*}[t]
\centering
\includegraphics[width=0.98\textwidth]{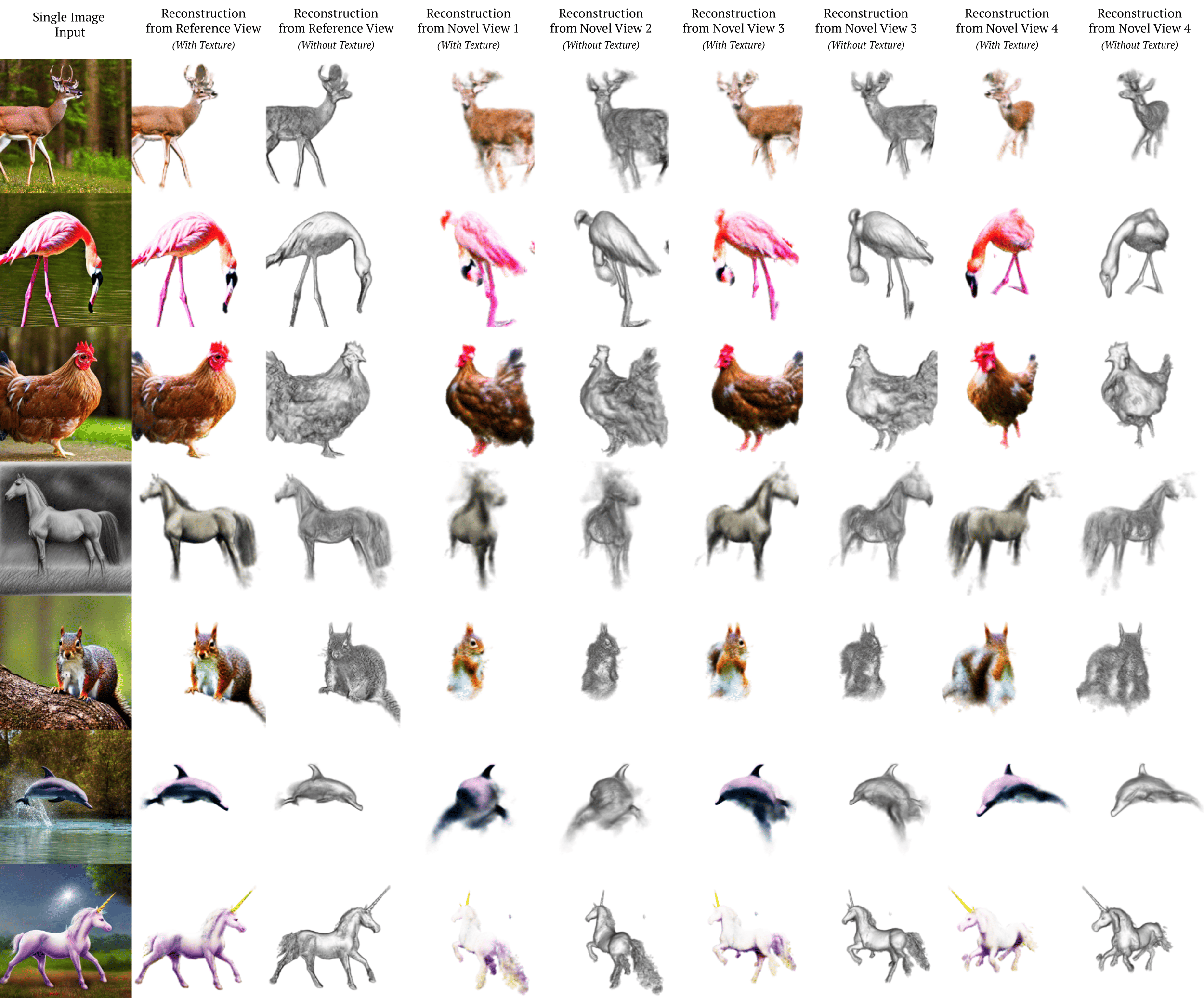}
\caption{
    \textbf{Text-to-Image-to-3D.} 
    This figure presents examples from our model using images generated directly from text prompts using Stable Diffusion~\cite{rombach22high-resolution}. The images were generated with the prompt ``An image of a \_\_\_\_'' where the blank space is replaced by ``deer'', ``flamingo'', ``hen'', ``pencil drawing of a horse'', ``squirrel'', ``dolphin'', and ``unicorn'', respectively. The results demonstrate that our method is able to reconstruct plausible object shapes even from synthetic images. 
}
\label{fig:supplementary_results_2_text2image}
\end{figure*}

\begin{figure*}[t]
\centering
\includegraphics[width=0.98\textwidth]{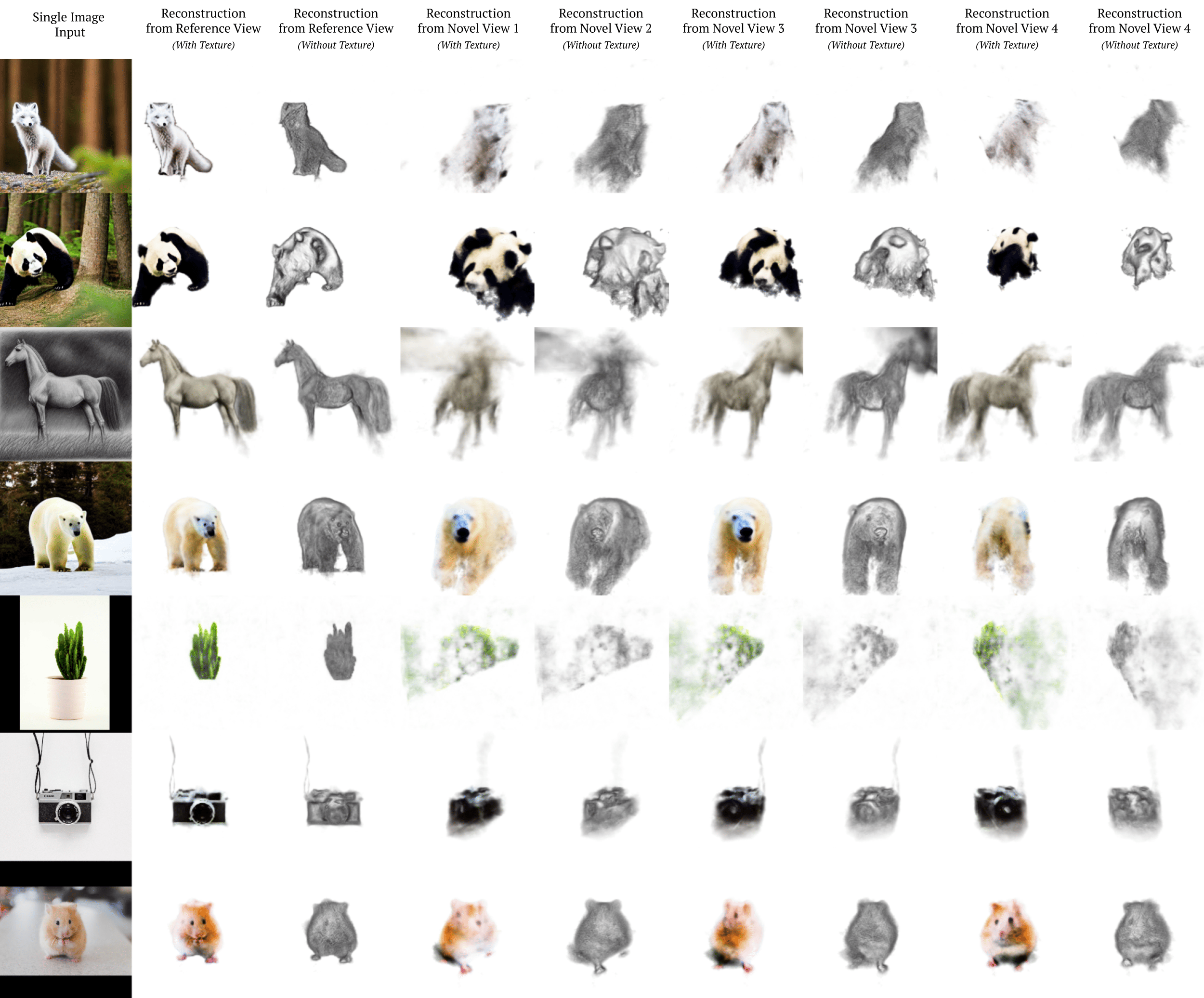}
\caption{
    \textbf{Additional failure cases.} 
    This figure presents additional failure cases from our model. The first column shows the input image. The second column shows the reconstruction from the reference viewpoint. The following columns show renders from novel viewpoints, which make clear why these examples are failure cases. Note that some examples (for example, the panda bear in the second row and the hamster in the last row) suffer from the Janus problem.
}
\label{fig:supplementary_results_3_failure}
\end{figure*}

Our use of view-specific prompting partially alleviates this issue. For example, when we render an image of a panda from the back, we optimize using the text prompt ``An image of a $\langle$object$\rangle$, back view'', where ``$\langle$object$\rangle$'' is our image-specific token corresponding to the image of a panda. However, even with view-specific prompting, this problem still occurs. This problem is visible with the panda in \cref{fig:supplementary_results_2_text2image} (row 2). We note that this problem is not unique to our method; it can also be seen with \cite{poole22dreamfusion:} (see Figure 9, last row). 

\section{Unsuccessful Experiments and Regularization Losses}

In the process of developing our method, we experimented with numerous ideas, losses, and regularization terms which were not included in our final method because they either did not improve reconstruction quality or did not improve it enough to justify their complexity. Here, we describe some of these ideas for the benefit of future researchers working on this problem.

\paragraph{Using DM for reconstruction loss.} One idea we tried involved using the diffusion model within our reconstruction objective as well as our prior objective. This involved a modified version of $\mathcal{L}_{\text{SDS}}$ in which we compared the noise predicted by the diffusion model for our noisy rendered image to the noise predicted by the diffusion model for a noisy version of our input image. We found that with this loss we were able to reconstruct the input image to a certain degree, but that we did not match the exact input image colors or textures.
\paragraph{Normals smoothing in 3D.} Our normals smoothing term operates in 2D, using normals rendered via the NeRF equation. We also tried different ways of smoothing normals in 3D. However, possibly due to our grid-based radiance field and/or our finite difference-based normals computation, we found that these regularization terms were all very noisy and harmful to reconstruction quality. 
\paragraph{Using monocular depth.} We tried incorporating monocular depth predictions into the pipeline, using pre-trained monocular depth networks such as MiDaS~\cite{Ranftl2019}. Specifically, we enforced that the depth rendered from the reference view matched the depth predicted by MiDaS for the input image. We found that this additional depth loss in most instances did not noticeably improve reconstruction quality and in some cases was harmful. Nonetheless, these results are not conclusive and future work could pursue other ways of integrating these components. 
\paragraph{Using LPIPS and SSIM reconstruction losses.} We tried using LPIPS~\cite{zhang2018perceptual} and SSIM losses in place of our L2 reconstruction loss. We found that LPIPS performed similarly to L2, but incurred additional computation and memory usage. We found that SSIM without either L2 and LPIPS resulted in worse reconstruction quality, but that it yielded fine results when combined with them. We did not include it in our final objective for the sake of simplicity.
\paragraph{Rendering at higher resolutions.} Since Stable Diffusion operates on images of resolution $512$px, it is conceivable that rendering at higher resolution would be benefitial with regard to the prior loss. However, we found no noticeable difference in quality when rendering at higher resolutions than $96$px or $128$px. For computational purposes, we used resolution $96$px for all experiments in the main paper.
\paragraph{Using DINO-based prior losses.} Similarly to the CLIP prior loss, one could imagine using other networks to encourage renders from novel views to be semantically similar to the input image. Due to the widespread success of the DINO~\cite{caron21emerging} models in unsupervised learning, we tried using DINO feature losses in addition to the Stable Diffusion prior loss. Specifically, for each image rendered from a novel view, we computed a DINO image embedding and maximized its cosine similarity with the DINO image embedding of the reference image. We found that this did not noticeably improve or degrade performance. For purposes of simplicity, we did not include it. 

\section{Links to Images for Qualitative Results}

For our qualitative results, we primarily use images from datasets such as Co3D. We also use a small number of images sourced directly from the web to show that our method works on uncurated web data. We provide links to all of these images on our \href{https://lukemelas.github.io/realfusion/links}{project website}.

\clearpage

\begin{figure*}[t]
\centering
\includegraphics[width=0.98\textwidth]{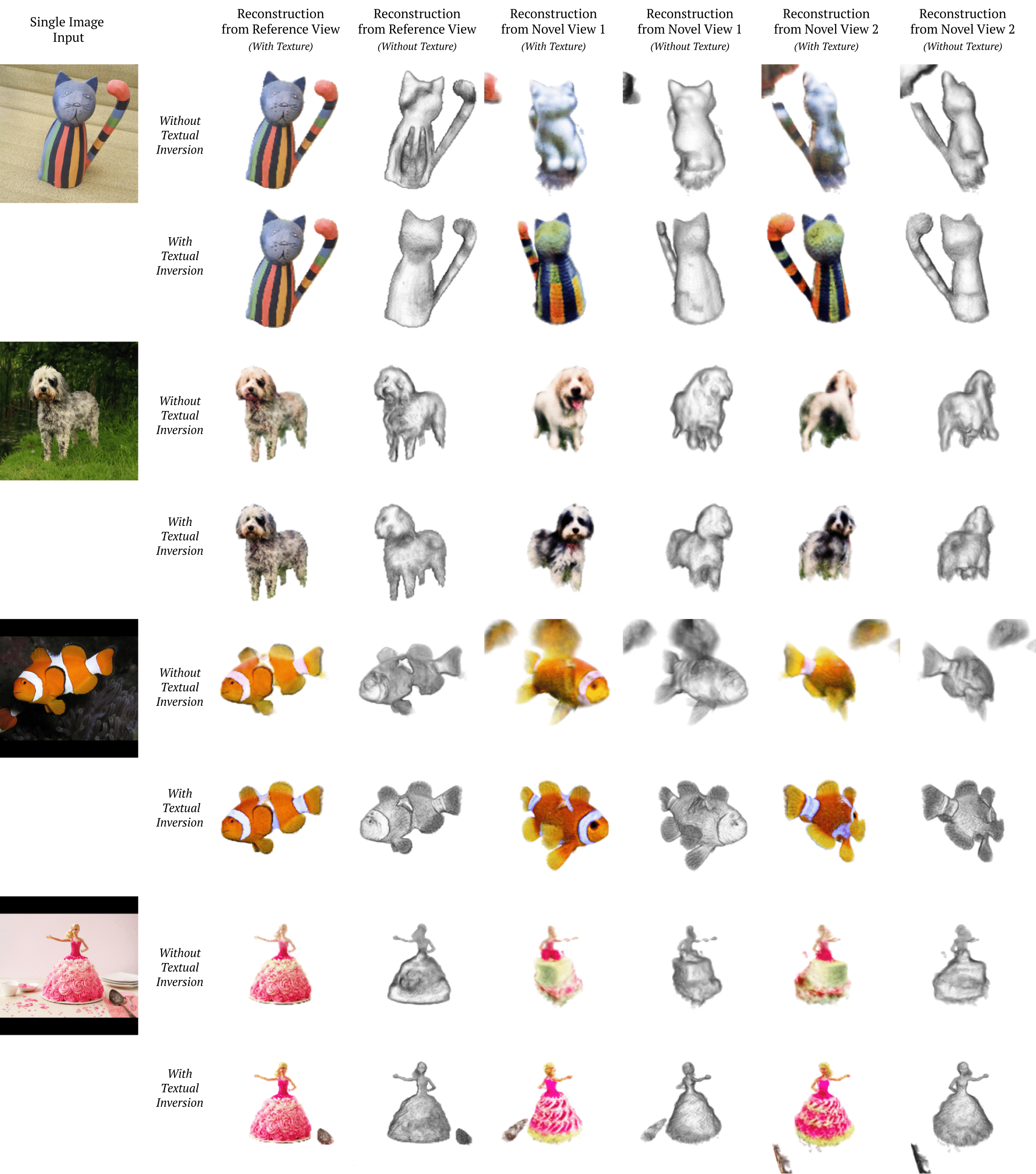}
\caption{
    \textbf{A visualization of the effect of single-image textual inversion on reconstruction quality.} An expanded version of Figure 7 in the main paper showing the effect of single-image textual inversion on reconstruction quality. The top row in each pair of rows shows reconstruction results using a standard text prompt, whereas the bottom row shows reconstruction results using single-image textual inversion. The novel views are chosen to show the back side of the object; note how the examples without textual inversion look like highly-generic versions of the objects in the input image. 
}
\label{fig:textual_inversion_supplementary}
\end{figure*}

\begin{figure*}[t]
\centering
\includegraphics[width=0.98\textwidth]{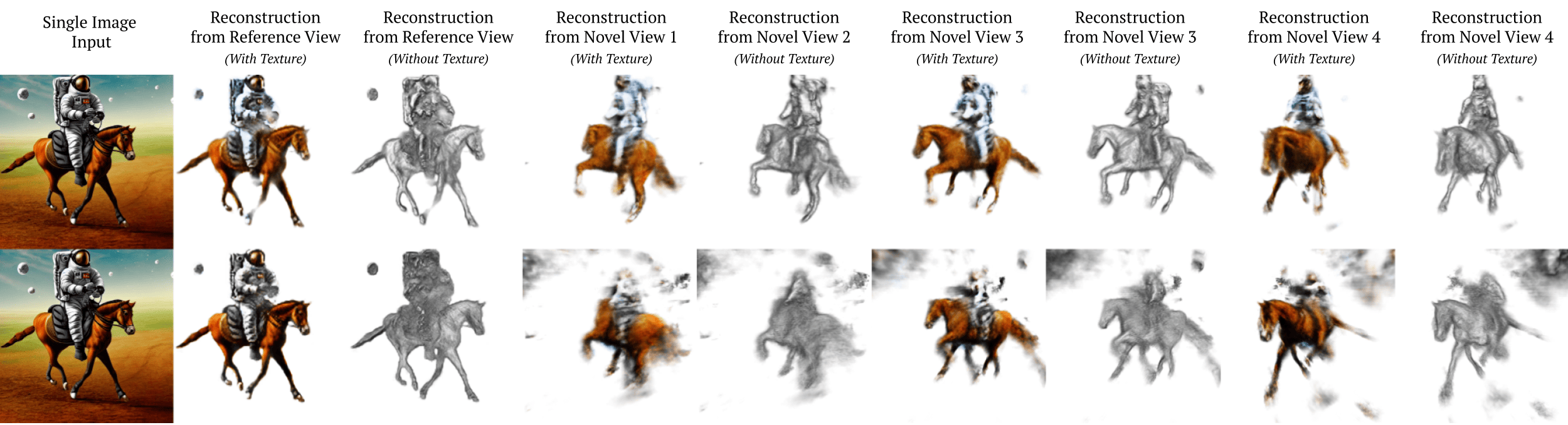}
\caption{
    \textbf{An example of variation across random seeds for a challenging input image.} As described in the main paper, our model is able to generate multiple reconstructions for a given input image. For this figure, we apply our method (in a text-to-image-to-3D manner) to a highly challenging image produced by Stable Diffusion from the text prompt ``An image of an astronaut riding a horse.'' We run reconstruction using two different seeds: one of these (top) yields a reasonable shape, whereas the other is a failure case that does not yield a reasonable shape. This example both highlights the ability of our method to reconstruct highly challenging shapes and also demonstrates how future work could aim to improve reconstruction consistency and quality. 
}
\label{fig:supplement_astronaut}
\end{figure*}

\end{document}